\documentclass[11pt]{article}

\usepackage[preprint]{acl}

\usepackage{times}
\usepackage{latexsym}

\usepackage[T1]{fontenc}

\usepackage[utf8]{inputenc}

\usepackage{microtype}

\usepackage{inconsolata}

\usepackage{graphicx}

\usepackage{booktabs}
\usepackage{multirow}
\usepackage[table]{xcolor}
\usepackage{makecell}

\title{CANDID: Candidate-Driven Intervention for Disciplined Execution in Local Computer-Use Agents}

\author{
  Woongkyu Lee \quad Jungwook Choi\thanks{Corresponding author.} \\
  Hanyang University \\ 
  \texttt{\{lwghanyang, choij\}@hanyang.ac.kr}
}

\begin{document}
\maketitle
\begin{abstract}
We present CANDID (Candidate-Driven Intervention for Disciplined Execution), an execution-control framework designed to enhance the reliability of local computer-use agents (CUAs) in resource-constrained settings. While local CUAs feature robust baseline proficiency in single-step action prediction, they frequently fail at critical junctions due to rigid action loops or premature DONE decisions. Rather than imposing imprecise cognitive loads through textual reflection, CANDID repurposes parallel test-time scaling into a disciplined, execution-level intervention at sparse, failure-prone points. We introduce a loop-recovery mechanism to filter candidates away from repeated behaviors and a conservative, unanimous DONE gating policy that requires absolute consensus to terminate. Experiments on OSWorld demonstrate that CANDID's structural control significantly improves success rates, outperforming feedback-based alternatives with minimal execution overhead.
\end{abstract}

\section{Introduction}
\label{sec:introduction}

Computer-use agents (CUAs) have recently emerged as a promising paradigm for autonomous interaction with graphical user interfaces (GUIs), enabling multimodal agents to execute real-world desktop and web tasks by understanding screenshots and executing GUI actions. Recent advances in multimodal foundation models~\cite{qin2025uitarspioneeringautomatedgui, wang2025opencuaopenfoundationscomputeruse, xue2026evocuaevolvingcomputeruse, xu2025aguvisunifiedpurevision, lin2024showuivisionlanguageactionmodelgui} and agentic frameworks~\cite{agashe2024agentsopenagentic, agashe2025agents2compositionalgeneralistspecialist, gonzalezpumariega2026scalingagentscomputeruse, yang2025gta1guitesttimescaling} have significantly improved long-horizon GUI interaction capabilities across increasingly realistic environments~\cite{xie2024osworldbenchmarkingmultimodalagents, zhou2024webarenarealisticwebenvironment, bonatti2024windowsagentarenaevaluating, deng2023mind2webgeneralistagentweb}. In particular, modern CUA systems leverage hierarchical planning~\cite{agashe2024agentsopenagentic}, memory retrieval~\cite{liu2025paluiplanningactivelookback}, and test-time scaling~\cite{yang2025gta1guitesttimescaling} strategies to improve execution robustness and task success rates.

Deploying CUAs under local, resource-constrained settings is practically important, and small language model (SLM)-based agentic systems have been increasingly proposed toward this goal~\cite{belcak2025smalllanguagemodelsfuture}. Yet such deployment remains highly challenging: unlike large cloud-based systems that rely on expensive multi-agent orchestration~\cite{song2026coact1computerusingmultiagentcoding, yang2026ossymphonyholisticframeworkrobust, han2026vlaaguiknowingstoprecover} or proprietary reasoning models~\cite{opus4.6, singh2026openaigpt5card}, local CUAs operate with limited model sizes and weaker multimodal reasoning. Under these conditions, simply increasing execution horizons or enlarging multimodal histories often fails to reliably improve performance. The central obstacle is that local agents fail in two coupled ways: they fall into repetitive action loops, or terminate prematurely before completing the task. Moreover, naively scaling the step budget or contextual history merely shifts the failure pattern between the two, rather than resolving both.

Prior methods for recovering failed agent trajectories primarily rely on textual reflection or verifier feedback~\cite{shinn2023reflexionlanguageagentsverbal, han2026vlaaguiknowingstoprecover}, where the agent inspects its own recent trajectory, generates natural-language feedback, and incorporates it before revising its next action. However, we argue that such feedback-based approaches provide limited benefit for lightweight local CUAs, whose reasoning capabilities are often insufficient either to generate meaningful textual feedback about the failure or to convert given feedback into distinct, executable GUI actions. These limitations prevent agents from escaping repetitive action loops or improving termination decisions under ambiguous completion states, often causing the model to still repeat its failed behavior under a restated rationale.

To overcome these limitations, we instead build on a key observation about local CUA behavior: although they struggle to assess trajectory progress, they remain capable of generating multiple plausible actions—which we refer to as \textit{candidates}—from the same interface state. This suggests that the bottleneck lies not in the availability of good candidates, but in committing to one when progress or completion is uncertain. We therefore propose CANDID, a candidate-based control framework that treats candidate disagreement as an execution-control signal, rather than asking the model to reflect on its own failure. CANDID instantiates this principle through two modules: (1) a loop-recovery module that identifies repetitive behavior through rule-based detection and selects an alternative candidate, and (2) a unanimous DONE gating module that accepts termination only when all sampled candidates agree. CANDID applies this sparse intervention only at these failure-prone points, concentrating computation where it is most beneficial while keeping overhead low.

We evaluate CANDID on the OSWorld benchmark~\cite{xie2024osworldbenchmarkingmultimodalagents} using multiple local CUA models, including Qwen3-VL~\cite{bai2025qwen3vltechnicalreport} and OpenCUA~\cite{wang2025opencuaopenfoundationscomputeruse} variants. Experimental results show that CANDID consistently improves task success rates across all evaluated models, outperforming both feedback-based interventions and a test-time scaling baseline. Further analyses show that its two control modules address complementary failure modes, and that sparse intervention is more effective than applying candidate selection uniformly throughout execution.

Our contributions are summarized as follows:
\begin{itemize}
\item We identify repetitive action loops and premature false success as dominant coupled failure modes in local CUAs that cannot be reliably resolved through naive execution scaling or textual feedback.
\item We propose CANDID, a sparse, candidate-based control framework that leverages multiple candidates generated by the same local agent at failure-prone steps, through \textit{loop-recovery candidate filtering} and \textit{unanimous} \texttt{DONE} \textit{gating}, without modifying the model.
\item We demonstrate that CANDID consistently improves task success across multiple local CUAs, outperforming feedback-based and test-time scaling baselines, and through targeted ablations we show that these gains stem specifically from CANDID's sparse, candidate-driven intervention design with minimal runtime overhead.
\end{itemize}

\section{Background and Motivation}
\label{sec:background}

\subsection{Computer-Use Agents}

\begin{figure}[t]
    \centering
    \includegraphics[width=\linewidth]{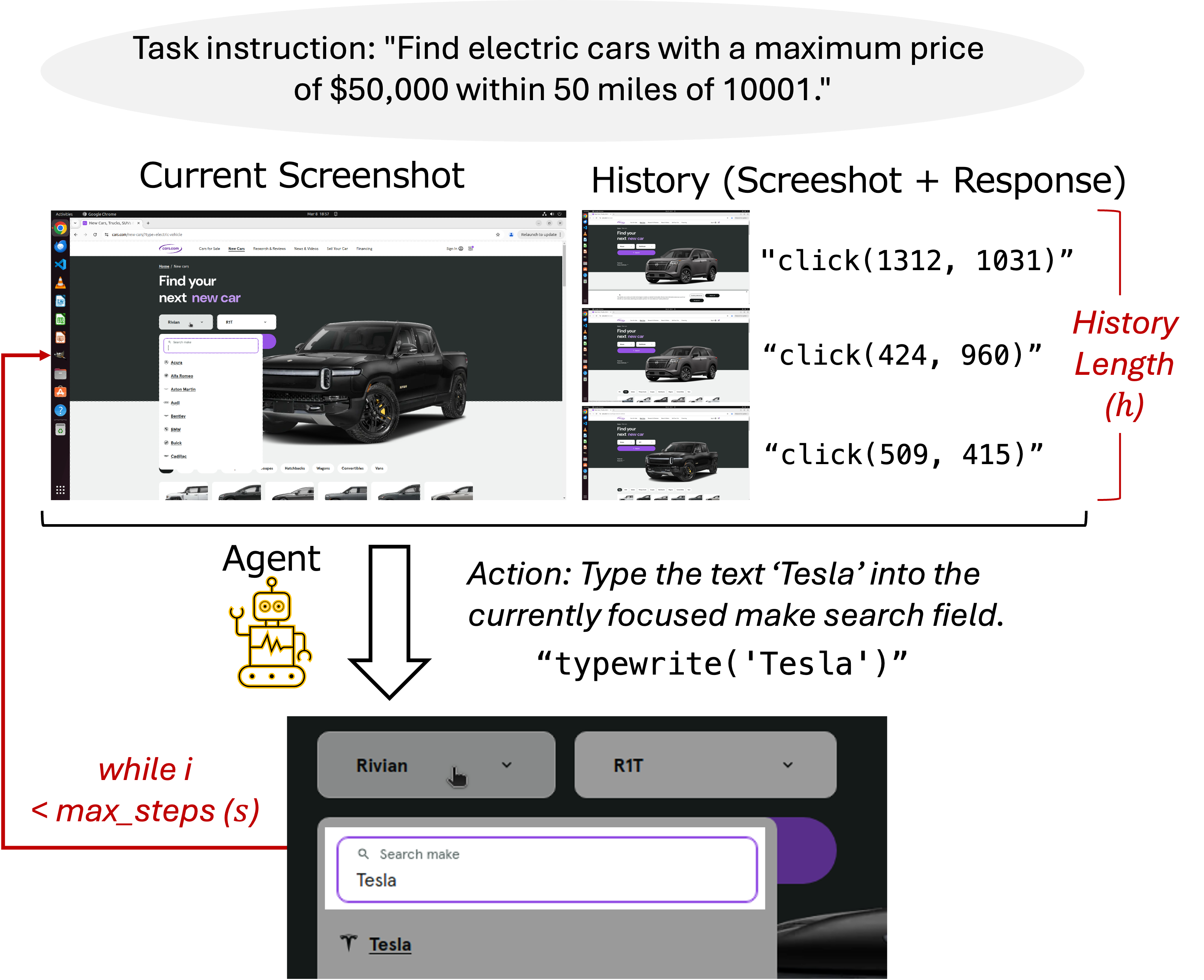}
    \caption{Operational workflow of a computer-use agent. The agent takes the task instruction and a multimodal history of $h$ past [screenshot, response] pairs as input to generate Python-executable GUI actions up to a maximum step budget ($s$).}
    \label{fig:cua_background}
\end{figure}

Computer-use agents (CUAs) autonomously operate graphical user interfaces (GUIs) in real OS environments, observing screenshots and generating executable actions such as clicking, typing, scrolling, or keyboard inputs. As illustrated in Fig.~\ref{fig:cua_background}, at each step the agent receives a task instruction, the current screenshot, and a multimodal history of previous screenshot--response pairs, and predicts the next action in a structured format (e.g., \texttt{click(x,y)} or \texttt{typewrite(`Tesla')}). The action is executed in the environment, after which the updated screenshot is observed. This interaction continues until the agent emits a termination signal such as \texttt{DONE} or reaches the step budget $s$, while the history length $h$ controls the number of past screenshot--response pairs retained as context.



\begin{figure}[t]
\centering
\includegraphics[width=\linewidth]{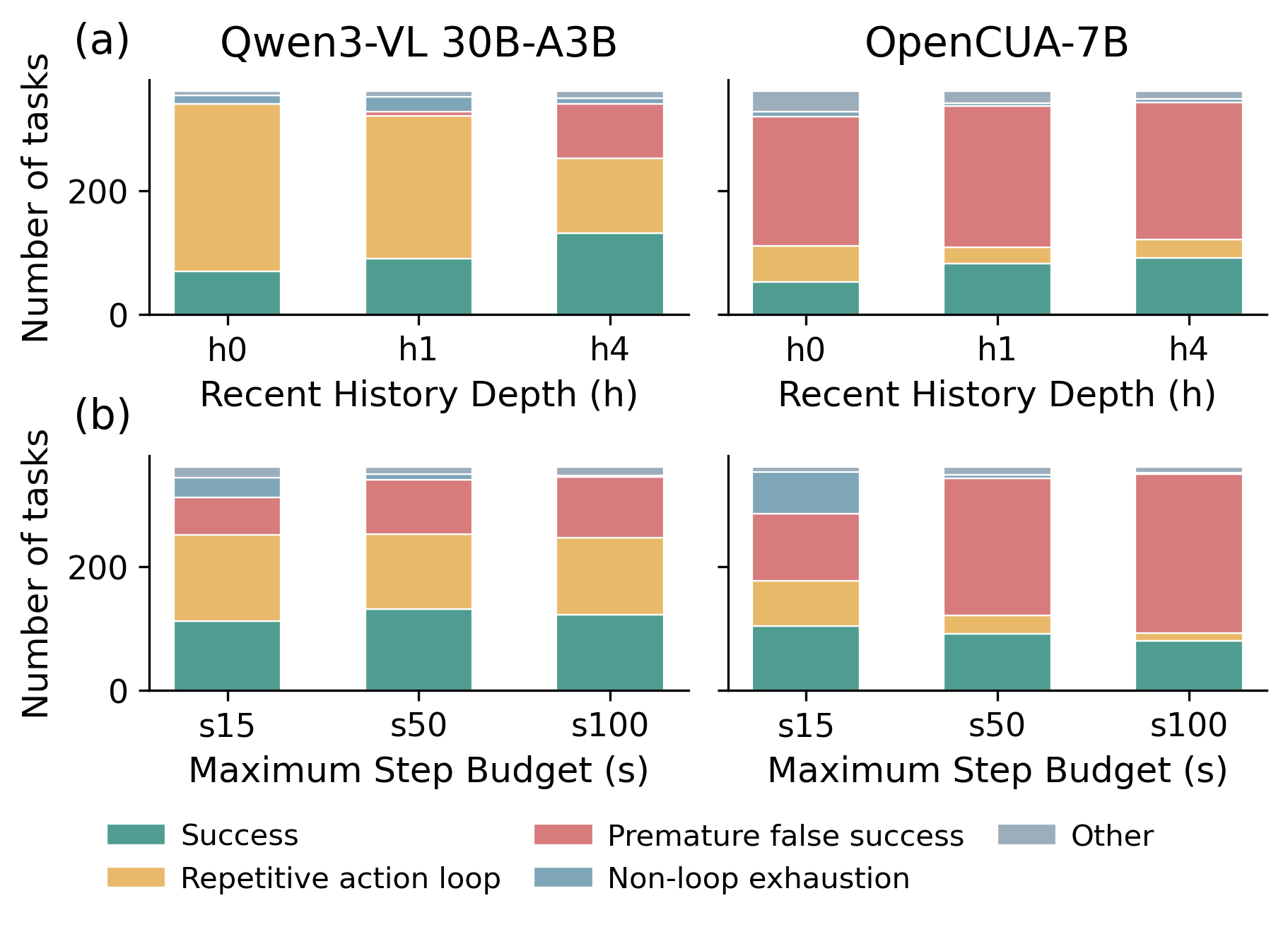}
\caption{
Failure modes of local CUAs on the OSWorld benchmark.
Stacked bars show task outcomes: success, repetitive action loop, premature false success, step budget exhaustion without loops, and other failures.
(a) Varying recent history depth ($h \in \{0,1,4\}$) with $s=50$.
(b) Varying maximum step budget ($s \in \{15,50,100\}$) with $h=4$.
}
\label{fig:failure_mode_analysis}
\end{figure}

\subsection{Failure Modes in Local CUAs}
\label{sec:analysis-failure-modes}

We analyze the behavioral limitations of local CUAs by evaluating Qwen3-VL-30B-A3B and OpenCUA-7B on OSWorld under varying step budgets $s \in \{15,50,100\}$ and history depths $h \in \{0,1,4\}$. As shown in Fig.~\ref{fig:failure_mode_analysis}, repetitive action loops and premature false success dominate across nearly all settings, while non-loop exhaustion remains limited except at the tightest budget ($s=15$). Qwen3-VL is more prone to repetitive loops, whereas OpenCUA more often terminates prematurely. Increasing the history depth or step budget reduces some exhaustion cases but shifts failures toward false completion rather than success. These results indicate that naive execution scaling does not resolve the dominant failure modes—it only trades one form of failure for another.

\subsection{Limitations of Textual Feedback}
\label{sec:analysis-feedback}

\begin{table}[t]
\centering
\small
\resizebox{\linewidth}{!}{%
\begin{tabular}{lcc}
\toprule
\textbf{Feedback} & \textbf{Metric} & \textbf{Rate (\%)} \\
\midrule
\multirow{3}{*}{Loop Breaker}
& Repeated exact action & 61.5 \\
& Repeated type/target & 64.3 \\
& Failure after recovery & 92.8 \\
\midrule
\multirow{3}{*}{Verifier}
& Repeated \texttt{DONE} after rejection & 73.4 \\
& Next action repeats recent target & 44.3 \\
& Failure after rejection & 75.9 \\
\bottomrule
\end{tabular}
}
\caption{Behavioral outcomes after feedback interventions on Qwen3-VL-30B-A3B. Despite receiving Loop Breaker or Verifier feedback, the agent frequently repeats the same behavior or continues to fail, indicating that feedback alone offers limited recovery capability.}
\label{tab:textual_feedback_behavior}
\end{table}

Prior autonomous-agent frameworks often use prompt-level verification or self-reflection to mitigate execution failures~\cite{han2026vlaaguiknowingstoprecover,pan2024autonomousevaluationrefinementdigital,shinn2023reflexionlanguageagentsverbal}. These methods work well with proprietary, large-scale reasoning models, whose stronger language understanding lets verbal feedback translate into corrective actions. Inspired by VLAA-GUI~\cite{han2026vlaaguiknowingstoprecover}, we implement two representative self-feedback mechanisms: a \textit{Loop Breaker}, which detects repetitive behavior and appends a textual recovery reflection, and a \textit{Verifier}, which rejects premature \texttt{DONE} predictions and provides corrective feedback.

However, such textual reflection offers limited recovery capability for smaller, local-scale CUA models, as verbal critiques often fail to induce a distinct executable action once the agent enters a localized loop or faces goal ambiguity. As shown in Table~\ref{tab:textual_feedback_behavior}, Loop Breaker feedback frequently leads the agent to repeat the same action or target and ultimately fail; for instance, after repeatedly attempting to enable a toggle switch, the reflection merely restates the same instruction rather than suggesting an alternative action. Similarly, after a rejected \texttt{DONE} prediction, the agent regenerates \texttt{DONE} or repeats a recent target, again resulting in failure. These results highlight the limited effectiveness of textual feedback for local CUAs.



\begin{figure}[t]
    \centering
    \includegraphics[width=\linewidth]{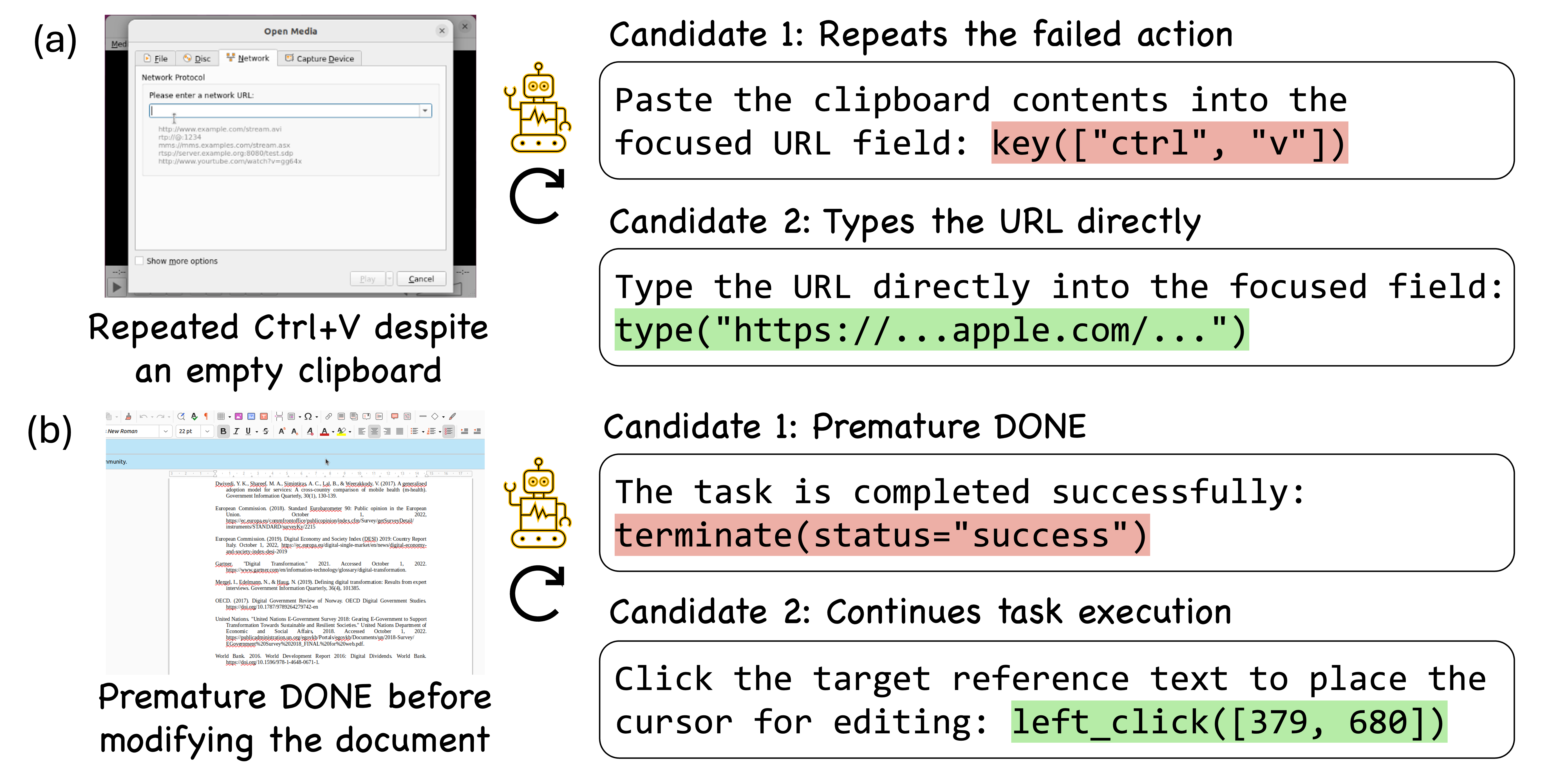}
    \caption{A local agent can generate valid alternative candidate actions at failure-prone states. In (a), the agent repeatedly presses \texttt{Ctrl+V} despite an empty clipboard, while another candidate types the URL directly. In (b), the agent prematurely declares task completion before modifying the document, while another candidate continues execution by clicking the target text for editing.}
    \label{fig:candidate-examples}
\end{figure}

\subsection{Generating Failure-Escaping Candidates}


\begin{figure*}[t]
    \centering
    \includegraphics[width=\textwidth]{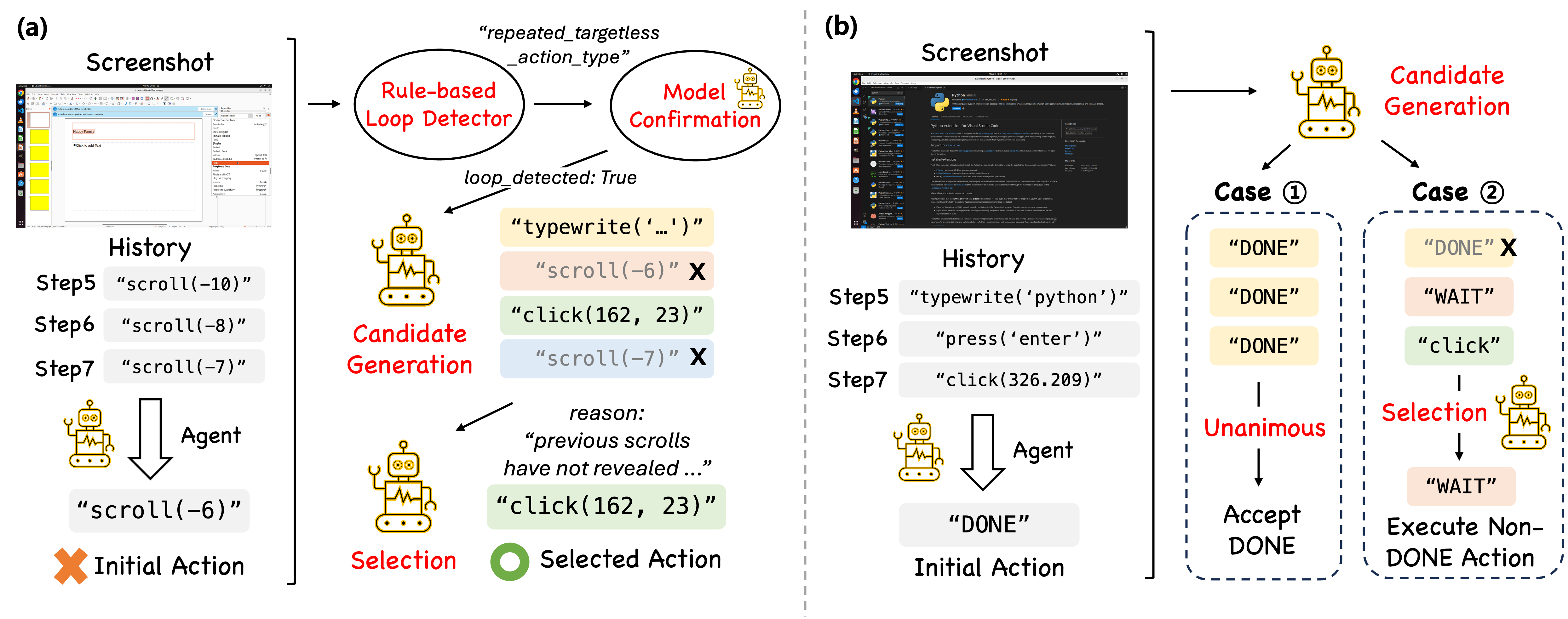}
    \caption{Overview of CANDID, our candidate-based execution-control framework for local computer-use agents. CANDID intervenes at two high-risk control points: (a) \textbf{Loop Recovery:} A loop detector monitors the execution history for repetitive local behavior and triggers structured candidate generation and filtering to escape the loop. (b) \textbf{Unanimous DONE Gating:} When the baseline agent proposes \texttt{DONE}, CANDID samples additional action candidates from the same context and executes \texttt{DONE} only when all valid candidates agree on termination.}
    \label{fig:method_overview}
\end{figure*}

Although local CUAs often remain trapped in repetitive loops or terminate prematurely, these failures do not necessarily indicate an inability to generate useful executable actions. As illustrated in Fig.~\ref{fig:candidate-examples}, sampling multiple actions from the same execution state can reveal executable alternatives that escape the current failure pattern: an agent that repeatedly presses \texttt{Ctrl+V} despite an empty clipboard may instead generate an action that types the URL directly, while an agent that prematurely declares completion may generate a continuation action that resumes document editing. This suggests that the bottleneck lies not in generating executable alternatives, but in committing to a single action despite disagreement among plausible candidates. This capability is not limited to isolated examples: when the agent exhibits repetitive-action behavior, all four sampled candidates are distinct at the exact-action level in 47.7\% of cases, whereas complete candidate collapse occurs in only 8.4\% (Appendix~\ref{sec:candidate-diversity}).

\section{Method}
\label{sec:method}

We propose CANDID, a sparse, candidate-based execution-control framework for local CUAs that intervenes at failure-prone steps to address two coupled failure modes: repetitive action loops and premature \texttt{DONE} decisions. Our key intuition is that local CUAs can often generate multiple plausible candidates from the same interface state but fail by committing to a single one under uncertain execution states. Instead of sampling throughout execution, CANDID generates a small set of candidates only at these high-risk points, recovering from loops by selecting an alternative candidate and preventing premature termination by requiring unanimous agreement before accepting \texttt{DONE}.

\subsection{Loop-Recovery Candidate Selection}
\label{sec:method-loop}


Fig.~\ref{fig:method_overview}(a) illustrates CANDID's loop-recovery pipeline. At each step, the agent predicts its next action from the current screenshot and recent history. When the proposed action is identified as a potential repetitive loop and confirmed as stagnant, CANDID samples alternative candidates, filters out those matching the detected loop pattern, and executes one of the remaining candidates.



\paragraph{Loop detection.}
Judging whether the agent is stuck in a loop is itself unreliable: the local CUA often cannot assess its own progress, while pure rule-based heuristics risk flagging legitimate repetition as a loop. We therefore use a two-stage gate that separates cheap detection from reliable confirmation. A lightweight rule-based detector first compares the proposed action against a sliding window of recent actions (governed by a window size, a minimum repeat count, and a target-coordinate tolerance) and flags repetitive patterns—for example, repeated scrolling is flagged as \texttt{repeated\_targetless\_action\_type}. Only for flagged cases does the local model then confirm whether the trajectory is genuinely stuck in a loop. This confines the model's harder judgment to the rare cases where it is needed, while leaving the initial action untouched otherwise. Further details and sensitivity analysis are in Appendix~\ref{sec:appendix-loop-detection} and Appendix~\ref{sec:loop-param}, respectively.



\paragraph{Recovery candidate generation.}
Simply requesting a single new action from the same context after a loop is confirmed often reproduces the same failed action. Instead, once a loop is confirmed, the local agent samples a small set of executable candidates (a hyperparameter set to four in our main experiments) from the current environment state, leveraging its demonstrated ability to produce diverse actions under repeated sampling. These candidates may include both repeated and alternative actions, and are passed to the filtering stage.

\paragraph{Filtering and selection.}
Before the agent selects among candidates, CANDID first narrows the pool to reduce selection overhead: malformed or non-executable actions are removed, \texttt{DONE} is disallowed during loop recovery, candidates matching the detected loop pattern are excluded, and duplicates are merged. If no valid candidate remains, CANDID falls back to the original proposal; if exactly one remains, it is executed directly; otherwise, the agent selects among the filtered candidates based on the task instruction and recent trajectory.



\subsection{Unanimous DONE Gating}
\label{sec:method-done}

Local CUAs frequently misjudge task completion, often terminating prematurely (Section~\ref{sec:analysis-failure-modes}), motivating a conservative approach to accepting \texttt{DONE}. Fig.~\ref{fig:method_overview}(b) illustrates CANDID's unanimous \texttt{DONE} gating module, which is triggered only when the agent proposes \texttt{DONE}. Rather than accepting a single termination prediction, CANDID samples additional candidates from the same context and accepts \texttt{DONE} only when all valid candidates agree (Case~1); even a single continuation candidate, such as \texttt{WAIT} or \texttt{click}, is treated as evidence that the task may still be unfinished, in which case termination is rejected and all \texttt{DONE} candidates are removed (Case~2). This conservative policy uses the agent's own candidate diversity to challenge unsupported \texttt{DONE} decisions without relying on an external verifier. We further show in Section~\ref{sec:ablation} that this unanimous policy more effectively suppresses premature completions than a majority voting.

\paragraph{Candidate selection.}
Once \texttt{DONE} candidates are removed, the remaining continuation candidates are processed for execution. In Case~2, for example, the sampled set contains \texttt{DONE}, \texttt{WAIT}, and \texttt{click}; after filtering, only \texttt{WAIT} and \texttt{click} remain as continuation candidates, and exact duplicates are merged. If a single continuation candidate remains, it is executed directly; otherwise, the local agent selects one from the remaining continuation candidates based on the task instruction, current screenshot, and recent trajectory. The selector is restricted to choosing among the generated candidates and cannot introduce a new action.

\subsection{Profiling-Based Auto-Configuration}

\begin{table}[t]
\centering
\small
\resizebox{\linewidth}{!}{%
\begin{tabular}{lccc}
\toprule
& \makecell{Qwen3-VL\\8B}
& \makecell{Qwen3-VL\\30B-A3B}
& OpenCUA-7B \\
\midrule
\makecell[l]{Exact-action\\loop rate}
    & 32.12\% & 29.79\% & 7.09\% \\
\midrule
Unanimity rate
    & 88.2\% & 78.2\% & 40.8\% \\
Avg. unique alts.
    & 1.23 & 1.45 & 2.34 \\
Multi-type events
    & 1.1\% & 4.3\% & 23.4\% \\
\midrule
Applied control
    & Loop & Loop + DONE & DONE \\
\bottomrule
\end{tabular}%
}
\caption{Model-specific diagnostics with $k{=}4$ resampled candidates and the resulting control configuration. Exact-action loop rate denotes the fraction of steps that repeat the preceding action. Avg. unique alts. denotes the average number of unique non-\texttt{DONE} alternatives in disagreement cases, while multi-type events denotes the fraction of disagreements spanning multiple coarse action types.}
\label{tab:done-diagnostics}
\end{table}

The two control modules target distinct failure modes and may benefit local CUAs differently. Rather than tuning configurations post hoc, we estimate each module's potential effectiveness using rollout diagnostics that are independent of task success, and apply a module only when its target condition is clearly met. As shown in Table~\ref{tab:done-diagnostics}, loop recovery targets models with high exact-action loop rates (Section~\ref{sec:analysis-failure-modes}): Qwen3-VL-8B and Qwen3-VL-30B-A3B show loop rates roughly four times higher than OpenCUA-7B (32.12\% and 29.79\% vs. 7.09\%) and are assigned loop recovery. Unanimous \texttt{DONE} gating instead requires diverse, semantically distinct alternatives on disagreement to have any effect; OpenCUA-7B shows both lower \texttt{DONE} unanimity (40.8\% vs. 78–88\%) and far more multi-type disagreements (23.4\% vs. 1.1–4.3\%), and Qwen3-VL-30B-A3B shows a smaller but non-trivial version of the same pattern, while Qwen3-VL-8B shows almost none. We therefore apply \texttt{DONE} gating to OpenCUA-7B and Qwen3-VL-30B-A3B, but not to Qwen3-VL-8B. These diagnostics are computed once before the main evaluation and fixed as a single configuration per model.

\section{Results}
\label{sec:results}

\begin{table*}[t]
\centering
\small
\resizebox{\textwidth}{!}{%
\begin{tabular}{llccccccc}
\toprule
\multirow{2}{*}{\textbf{Model}} & \multirow{2}{*}{\textbf{Method}} & \multirow{2}{*}{\textbf{Trials}} & \multicolumn{5}{c}{Domain Accuracy (\%)} & \multirow{2}{*}{\textbf{Overall (\%)}} \\
\cmidrule(lr){4-8}
& & & \textbf{Office} (117) & \textbf{Daily} (78) & \textbf{Prof.} (49) & \textbf{Workflow} (93) & \textbf{OS} (24) & \\
\midrule
\multirow{2}{*}{\makecell[l]{Qwen3-VL-30B-A3B\\(s=50)}}
& Baseline & 5 & 37.43 & 48.15 & 41.63 & 18.25 & 36.67 & 35.31 $\pm$ 0.65 \\
& CANDID & 5 & 39.42 & 49.76 & 46.94 & 22.52 & 46.67 & \textbf{38.80 $\pm$ 1.24} {\color{gray}(+3.49)} \\
\midrule

\multirow{2}{*}{\makecell[l]{Qwen3-VL-30B-A3B\\(s=100)}}
& Baseline & 5 & 37.54 & 48.23 & 41.23 & 18.78 & 36.16 & 35.42 $\pm$ 1.34 \\
& CANDID & 5 & 40.63 & 50.59 & 48.57 & 23.82 & 42.93 & \textbf{39.68 $\pm$ 0.77} {\color{gray}(+4.26)} \\
\midrule

\multirow{2}{*}{\makecell[l]{Qwen3-VL-8B\\(s=50)}}
& Baseline & 5 & 32.74 & 43.37 & 44.49 & 14.77 & 44.57 & 32.82 $\pm$ 0.98 \\
& CANDID & 5 & 32.58 & 46.97 & 47.35 & 17.32 & 48.33 & \textbf{34.84 $\pm$ 1.27} {\color{gray}(+2.02)} \\
\midrule

\multirow{2}{*}{\makecell[l]{Qwen3-VL-8B\\(s=100)}}
& Baseline & 5 & 33.77 & 44.17 & 45.72 & 14.89 & 42.50 & 33.41 $\pm$ 1.29 \\
& CANDID & 5 & 34.45 & 48.39 & 43.45 & 17.63 & 50.00 & \textbf{35.42 $\pm$ 1.76} {\color{gray}(+2.01)} \\
\midrule

\multirow{2}{*}{\makecell[l]{OpenCUA-7B\\(s=50)}}
& Baseline & 5 & 23.92 & 36.31 & 44.90 & 11.39 & 39.13 & 27.21 $\pm$ 1.70 \\
& CANDID & 5 & 24.91 & 38.39 & 62.04 & 10.67 & 53.44 & \textbf{31.05 $\pm$ 0.97} {\color{gray}(+3.84)} \\
\midrule

\multirow{2}{*}{\makecell[l]{OpenCUA-7B\\(s=100)}}
& Baseline & 5 & 23.42 & 33.46 & 40.82 & 11.25 & 42.25 & 26.06 $\pm$ 2.16 \\
& CANDID & 5 & 25.36 & 40.20 & 56.73 & 10.93 & 47.83 & \textbf{30.60 $\pm$ 2.22} {\color{gray}(+4.54)} \\
\bottomrule
\end{tabular}
}
\caption{Main evaluation results on the OSWorld benchmark, averaged over multiple trials (trial counts noted per row). Values in parentheses next to domain names indicate the task count. Overall accuracy is reported as mean $\pm$ standard deviation; values in gray parentheses represent the absolute improvement over each model's baseline.}
\label{tab:main-results}
\end{table*}

\subsection{Setup}

\paragraph{Benchmark and Evaluation Setting.}
We evaluate our framework primarily on OSWorld~\cite{xie2024osworldbenchmarkingmultimodalagents}, excluding eight Google Drive-related tasks and reporting results on 361 tasks across five domains. We additionally assess cross-benchmark generalization on WindowsAgentArena~\cite{bonatti2024windowsagentarenaevaluating} with 154 tasks, with detailed settings and results provided in Appendix~\ref{sec:waa-results}. All agents use screenshot observations, generate PyAutoGUI-style actions, and use four recent screenshots as history ($h=4$) for the main experiments.


\paragraph{Models.}
Our evaluation covers three local computer-use agents: Qwen3-VL-30B-A3B-Instruct, Qwen3-VL-8B-Instruct, and OpenCUA-7B. This model selection encompasses both general-purpose vision-language models adapted to desktop control and models specialized for computer-use behavior, allowing us to test the proposed control mechanisms across different model scales and agent families.

\paragraph{Candidate sampling configuration.}
For both loop recovery and unanimous \texttt{DONE} selection, we fix the number of additional sampled candidates to four.
For Qwen-based agents, we use a temperature of 0.7 for default action generation and \texttt{DONE} candidate sampling, and a higher temperature of 1.0 for loop-recovery candidate generation to encourage greater action diversity under repetitive execution patterns.
For OpenCUA-7B, the default action generation uses greedy decoding (temperature 0.0), while candidate generation uses a temperature of 0.7 for both loop recovery and \texttt{DONE} selection.

\paragraph{Inference-Time Control Baselines.}
We compare CANDID against several inference-time control baselines. For test-time scaling, we apply GTA-1~\cite{yang2025gta1guitesttimescaling}, using the local CUAs as the planner while retaining the original GTA-1-7B grounding model and sampling four plans. We follow the publicly released GTA-1 implementation, including its available prompting and inference procedure, without modifying the disclosed components other than replacing the planner with the evaluated local CUA. We also implement step-wise \textit{Self-Guidance}, which generates compact progress feedback after each step and appends it to the next action-generation prompt. Finally, we implement VLAA-GUI-style~\cite{han2026vlaaguiknowingstoprecover} \textit{Loop Breaker} and \textit{Verifier} modules by adapting their core intervention concepts for local CUA evaluation, rather than reproducing the full multi-agent architecture. Additional details are provided in Appendix~\ref{sec:appendix-feedback-baselines}.



\subsection{Main Result}
\label{sec:main-results}

\paragraph{Overall performance.}
Table~\ref{tab:main-results} reports the main results on OSWorld. We evaluate CANDID under different step budgets to examine its effectiveness across varying execution lengths. Across all evaluated local CUAs, CANDID consistently improves task success over the corresponding baseline, yielding gains of 2.01--4.54 percentage points depending on the model and step budget. The improvements are observed under both shorter ($s=50$) and longer ($s=100$) budgets, demonstrating that CANDID remains effective across different interaction horizons.


\begin{table*}[t]
\centering
\small
\begin{tabular}{lcccccc}
\toprule
\multirow{2}{*}{\textbf{Method}} & \multicolumn{5}{c}{Domain Accuracy (\%)} & \multirow{2}{*}{\textbf{Overall (\%)}} \\
\cmidrule(lr){2-6}
& \textbf{Office} (117) & \textbf{Daily} (78) & \textbf{Professional} (49) & \textbf{Workflow} (93) & \textbf{OS} (24) & \\ 
\midrule
Baseline & 37.43 & 48.15 & 41.63 & 18.25 & 36.67 & 35.31 \\
\midrule
\rowcolor[HTML]{F9F9F9} \multicolumn{7}{l}{\textit{Test-Time Scaling Method}} \\
GTA-1 & 12.74 & 38.41 & 48.98 & 9.77 & 54.17 & 25.19 \\
\midrule
\rowcolor[HTML]{F9F9F9} \multicolumn{7}{l}{\textit{Feedback-Based Methods}} \\
Self-Guidance & 34.96 & 36.36 & 44.90 & 18.26 & 50.00 & 33.30 \\
VLAA-Loop Breaker & 37.75 & 46.65 & 46.94 & 21.36 & 42.12 & 36.98 \\
VLAA-Verifier & 36.39 & 50.22 & 40.14 & 14.74 & 34.72 & 34.22 \\
VLAA-Combined & 34.98 & 49.73 & 44.22 & 20.32 & 37.50 & 35.84 \\
\midrule
\textbf{CANDID} & 39.42 & 49.76 & 46.94 & 22.52 & 46.67 & \textbf{38.80} \\
\bottomrule
\end{tabular}
\caption{Comparison with inference-time control baselines using the Qwen3-VL-30B-A3B-Instruct. We compare CANDID against a GTA-1~\cite{yang2025gta1guitesttimescaling} test-time scaling method, step-wise Self-Guidance, and VLAA-GUI-style~\cite{han2026vlaaguiknowingstoprecover} feedback methods, including Loop Breaker, Verifier, and their combined variant. VLAA-GUI Loop Breaker is averaged over four runs, Verifier and Combined over three runs, and Baseline and CANDID over five runs; GTA-1 and Self-Guidance are based on a single run.}
\label{tab:feedback-comparison}
\end{table*}


\paragraph{Comparison with Inference-Time Control Methods.}
As shown in Table~\ref{tab:feedback-comparison}, CANDID outperforms the baseline and all inference-time control methods. Among the feedback-based methods, \textit{Self-Guidance} decreases accuracy from 35.31\% to 33.30\%. The VLAA-GUI-style \textit{Loop Breaker} provides a modest improvement to 36.98\%, whereas the \textit{Verifier} and the combined \textit{Loop Breaker+Verifier} variant achieve 34.22\% and 35.84\%, respectively. GTA-1 decreases overall accuracy to 25.19\%, suggesting limitations in the local CUA's high-level planning and structured plan generation. Overall, these results suggest that verbal feedback or planning-time sampling alone is insufficient for local CUAs, whereas CANDID improves execution stability by introducing alternative actions selectively at high-risk states.

\subsection{Ablation and Design Analysis}
\label{sec:ablation}



\begin{figure}[t]
    \centering
    \includegraphics[width=0.9\linewidth]{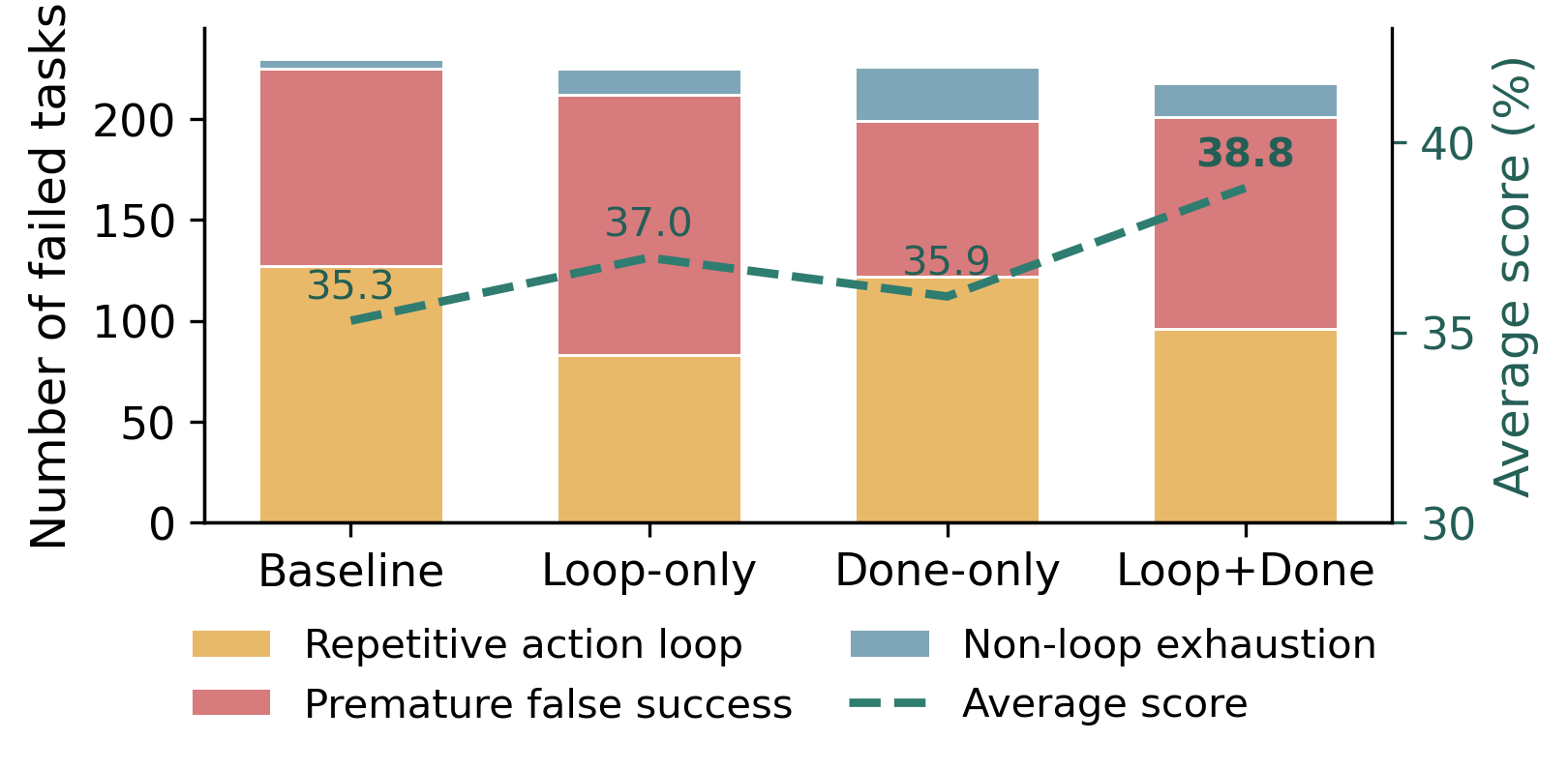}
    \caption{Failure-mode distribution and accuracy across component configurations on Qwen3-VL-30B-A3B ($s=50$). Loop Recovery primarily reduces repetitive loops, Unanimous DONE reduces premature false success, and their combination achieves the highest average score.}
    \label{fig:component_failure_modes}
\end{figure}

\paragraph{Complementary effects of control modules.}
Fig.~\ref{fig:component_failure_modes} shows that the two control modules affect distinct failure modes. Loop-only substantially reduces repetitive action loops but shifts some failures toward premature false success, while Done-only suppresses premature false success at the cost of increased execution exhaustion. Combining both modules mitigates this trade-off and achieves the highest average score of 38.8\%, demonstrating that loop recovery and conservative termination control are most effective when applied jointly.

\begin{table}[t]
\centering
\small
\resizebox{\linewidth}{!}{%
\begin{tabular}{lrrr}
\toprule
 & Baseline & Majority & Unanimous \\
\midrule
Accuracy (\%)     & 26.06  & 27.24  & 30.60 \\
Hit Max           & 19.33  & 29.33  & 64.67 \\
False Success     & 241.33 & 232.00 & 171.67 \\
Redirection (\%)  & --     & 5.53   & 49.93 \\
\bottomrule
\end{tabular}
}
\caption{Comparison of majority and unanimous \texttt{DONE} selection on OpenCUA-7B with a 100-step budget.}
\label{tab:majority_unanimous}
\end{table}


\paragraph{Majority vs. Unanimous DONE.}
Table~\ref{tab:majority_unanimous} compares majority voting with the proposed unanimous termination policy. Majority voting provides only a marginal improvement over the baseline, with a redirection rate of 5.53\%. In contrast, unanimous selection redirects 49.93\% of proposed terminations and reduces false success from 241.33 to 171.67, improving task accuracy from 26.06\% to 30.60\%. Although this increases max-step failures, the overall result indicates that majority agreement is too weak to counter the model's termination bias.

While unanimous gating more effectively suppresses premature termination, it does not constitute a reliable semantic completion verifier. In the task-level proxy analysis, it achieves 56.7\% precision and 26.3\% specificity, compared with 54.3\% and 15.2\% for the VLAA-GUI-style Verifier, respectively (Appendix~\ref{sec:termination-gate}). Thus, CANDID primarily acts as a conservative filter against unsupported termination rather than solving completion verification itself.

\begin{table}[t]
\centering
\small
\resizebox{\linewidth}{!}{%
\begin{tabular}{lrrr}
\toprule
 & Baseline & Per-Step Sampling & CANDID \\
\midrule
Accuracy (\%)   & 35.31 & 23.48 & 38.80 \\
Hit Max         & 135.0 & 239.0 & 116.6 \\
False Success   & 97.4  & 44.0  & 105.2 \\
Avg. Steps      & 25.61 & 36.89 & 24.40 \\
\bottomrule
\end{tabular}
}
\caption{Comparison of dense per-step sampling and sparse intervention with CANDID on Qwen3-VL-30B-A3B with a 50-step budget. Hit Max and False Success are non-mutually-exclusive trajectory statistics; a task may satisfy both conditions.}
\label{tab:dense_sparse}
\end{table}

\paragraph{Per-Step vs. Triggered Sampling.}
Table~\ref{tab:dense_sparse} compares CANDID with candidate sampling at every step. Per-step sampling lowers false success but substantially increases max-step failures and reduces accuracy to 23.48\%. In contrast, CANDID achieves 38.80\% accuracy while maintaining a similar trajectory length to the baseline. These results show that candidate sampling is effective when applied selectively at failure-prone decision points rather than throughout execution.

\begin{figure}[t]
    \centering
    \includegraphics[width=\linewidth]{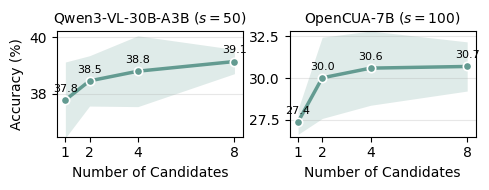}
    \caption{Effect of candidate count on task accuracy across the two local CUAs. Results with four candidates are averaged over five runs, while the remaining settings are averaged over three runs. Shaded regions denote $\pm1$ standard deviation across repeated runs.}
    \label{fig:candidate_sensitivity}
\end{figure}

\paragraph{Sensitivity to Candidate Count.}
Fig.~\ref{fig:candidate_sensitivity} examines the effect of the number of sampled candidates. Increasing the candidate count from one to four consistently improves task accuracy for both models. Further increasing it to eight provides only marginal gains, indicating diminishing returns beyond four candidates. We therefore use four candidates in the main experiments to balance performance and cost.

\begin{table}[t]
\centering
\small
\resizebox{\linewidth}{!}{%
\begin{tabular}{lcc}
\toprule
\textbf{Statistic} & \textbf{Loop Recovery} & \textbf{DONE Gating} \\
\midrule
Total Steps & 9371 & 9371 \\
Triggered & 2496 (26.6\%) & 429 (4.6\%) \\
Intervention Applied & 1426 (57.1\%) & 196 (45.7\%) \\
Original Action Kept & 1070 (42.9\%) & 233 (54.3\%) \\
Selector Invoked & 1043 (41.8\%) & 96 (22.4\%) \\
Selector Bypassed & 1453 (58.2\%) & 333 (77.6\%) \\
\bottomrule
\end{tabular}
}
\caption{Operational statistics of the candidate-based control framework on Qwen3-VL-30B-A3B. Percentages for ``Triggered'' are computed over all execution steps, while the remaining percentages are computed over the triggered steps.}
\label{tab:operation}
\end{table}


\subsection{Efficiency Analysis}
\label{sec:efficiency-analysis}

\paragraph{Operational statistics.}
Table~\ref{tab:operation} shows that CANDID intervenes selectively rather than modifying the entire execution process. Loop recovery is triggered in 26.6\% of steps, with 57.1\% resulting in trajectory changes, indicating that repetitive behaviors are frequent yet recoverable. \texttt{DONE} selection is triggered more sparsely, in only 4.6\% of steps, but redirects 45.7\% of those cases to non-\texttt{DONE} continuations. The selector is skipped in most cases, 58.2\% for loop recovery and 77.6\% for \texttt{DONE} selection, showing that many high-risk decisions can be resolved through structural filtering alone.

\begin{figure}[t]
    \centering
    \includegraphics[width=\linewidth]{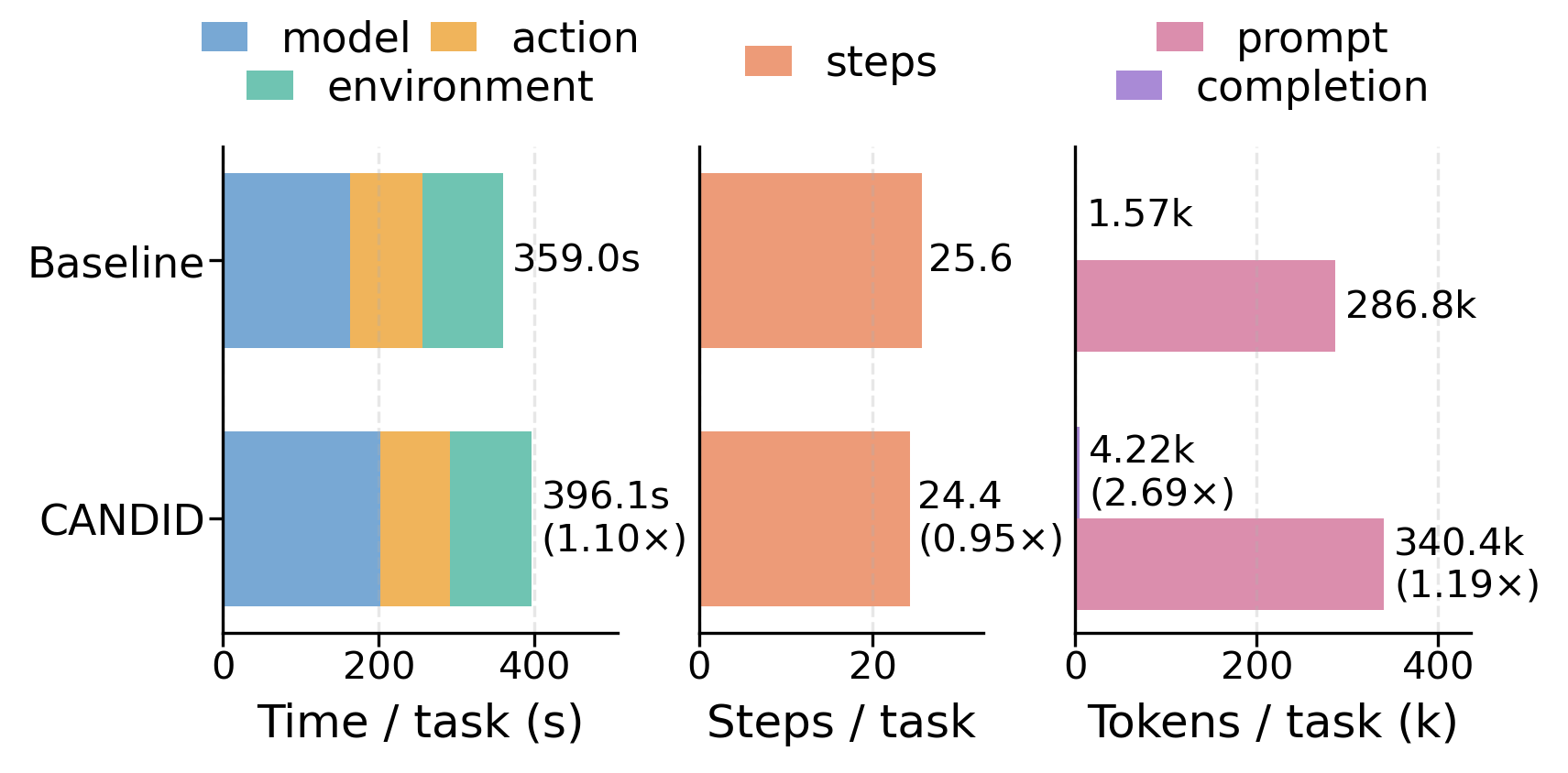}
    \caption{Efficiency comparison of baseline and CANDID on Qwen3-VL-30B-A3B. Execution time includes model inference, action execution, and environment initialization, with average steps and token usage.}
    \label{fig:efficiency}
\end{figure}

\paragraph{Runtime and token overhead.}
Fig.~\ref{fig:efficiency} compares the computational cost of CANDID and the baseline using Qwen3-VL-30B-A3B. CANDID increases prompt-token usage by only 19\%, while completion tokens, despite a larger relative increase from candidate generation and selection, remain negligible compared with prompt tokens. CANDID also shortens the average trajectory from 25.6 to 24.4 steps per task, partially offsetting the additional inference cost. Consequently, total execution time increases by only 10\% (359.0$\rightarrow$396.1 s/task), indicating that CANDID improves task performance with modest additional cost.

\subsection{Case Study}

Qualitative examples further illustrate how the two controls intervene at distinct failure points. Loop recovery redirects repeated \texttt{Enter} actions on a license screen to an executable alternative, \texttt{typewrite('n')}, while the DONE gating prevents premature termination before sending a prepared email by selecting a concrete continuation action. These cases show that CANDID converts detected risks into constrained executable alternatives. Additional details are provided in Appendix~\ref{sec:appendix-case-study}.

\section{Related Work}
\paragraph{Models and Frameworks.}

Recent progress in multimodal GUI agents has been driven by realistic computer-use benchmarks~\cite{xie2024osworldbenchmarkingmultimodalagents, deng2023mind2webgeneralistagentweb, rawles2025androidworlddynamicbenchmarkingenvironment} and CUA-specialized models that improve GUI grounding, action prediction, and long-horizon interaction~\cite{wang2025opencuaopenfoundationscomputeruse, qin2025uitarspioneeringautomatedgui, xue2026evocuaevolvingcomputeruse, lin2024showuivisionlanguageactionmodelgui, xu2025aguvisunifiedpurevision}. Agentic frameworks further improve reliability through multi-agent coordination~\cite{song2026coact1computerusingmultiagentcoding, yang2026ossymphonyholisticframeworkrobust}, hierarchical planning~\cite{agashe2025agents2compositionalgeneralistspecialist}, and test-time scaling~\cite{yang2025gta1guitesttimescaling, gonzalezpumariega2026scalingagentscomputeruse}. However, these approaches often rely on proprietary models or substantial inference-time computation, leaving efficient execution control for local CUAs comparatively underexplored.



\paragraph{Inference-Time Exploration and Selection.}
Reasoning-oriented LLM studies use parallel inference-time scaling through candidate sampling and branching search~\cite{wang2023selfconsistencyimproveschainthought, yao2023treethoughtsdeliberateproblem, zhu2025scalingtesttimecomputellm, lee2026agenticaggregationparallelscaling}, and related CUA frameworks adopt trajectory reranking or parallel exploration~\cite{yang2025gta1guitesttimescaling, gonzalezpumariega2026scalingagentscomputeruse}. While related in spirit, our approach applies candidate diversification only at failure-prone points to recover in local CUAs.

\paragraph{Execution Reliability and Recovery.}
Agent reliability has been enhanced via feedback, reflection, and trajectory refinement~\cite{madaan2023selfrefineiterativerefinementselffeedback, shinn2023reflexionlanguageagentsverbal, lin2026uivoyagerselfevolvingguiagent, han2026vlaaguiknowingstoprecover}. While VLAA-GUI improves automation using verifier and loop-breaker~\cite{han2026vlaaguiknowingstoprecover}, it depends on proprietary models~\cite{opus4.6, sonnet4.6, gemini31pro, singh2026openaigpt5card} and multi-agent systems. Conversely, CANDID improves reliability using only the local agent's own action candidates, without auxiliary agents or verifiers.

\section{Conclusion}
\label{sec:conclusion}


We presented CANDID, a candidate-based execution-control framework for local computer-use agents that addresses two coupled failure modes: repetitive action loops and premature DONE decisions, both of which persist under naive execution scaling and textual feedback. By leveraging alternative actions generated by the local agent and intervening only at failure-prone decision points, CANDID recovers from repetitive behavior and challenges unsupported termination. Experiments on OSWorld demonstrate consistent improvements across multiple local CUAs over feedback-based alternatives, with minimal execution overhead. These results suggest that selectively allocating inference-time computation to high-risk execution states can be more effective than uniformly scaling computation throughout the trajectory.



\section*{Acknowledgments}

This work was supported by the National Research Foundation of Korea (NRF) grant funded by the Korea government (MSIT) (No. RS-2025-00561961). It was also supported by the Institute of Information \& Communications Technology Planning \& Evaluation (IITP) grant funded by the Korea government (MSIT) (No. RS-2025-02214497, Development of low-level optimization program API technology for AI semiconductors), as well as by the Institute of Information \& Communications Technology Planning \& Evaluation (IITP) grant funded by the Korea government (MSIT) (No. RS-2026-25507427, Development of Efficient Architectures and Training Techniques for High-Performance Lightweight AI Models).

\section*{Limitations}

Our framework focuses on lightweight execution-time control for local computer-use agents and is evaluated primarily on OSWorld desktop environments. While the proposed mechanisms consistently improve robustness across multiple local CUAs, the current study does not explore broader settings such as mobile environments, large-scale cloud agents, or long-horizon multi-agent orchestration. CANDID also depends on the base agent producing sufficiently diverse and useful candidates; when sampled candidates collapse to the same failed behavior, candidate-based recovery provides limited benefit. Moreover, the current loop detector targets relatively local repetitive patterns and may miss longer-range or irregular failure cycles. Our candidate selection strategy relies on lightweight heuristic filtering and local reranking, and we do not yet integrate CANDID with higher-level planning or memory modules. Extending the framework with adaptive triggering, stronger verification, and higher-level agentic signals remains an important direction for future work.


\bibliography{reference}

@misc{han2026vlaaguiknowingstoprecover,
      title={VLAA-GUI: Knowing When to Stop, Recover, and Search, A Modular Framework for GUI Automation}, 
      author={Qijun Han and Haoqin Tu and Zijun Wang and Haoyue Dai and Yiyang Zhou and Nancy Lau and Alvaro A. Cardenas and Yuhui Xu and Ran Xu and Caiming Xiong and Zeyu Zheng and Huaxiu Yao and Yuyin Zhou and Cihang Xie},
      year={2026},
      eprint={2604.21375},
      archivePrefix={arXiv},
      primaryClass={cs.CL},
      url={https://arxiv.org/abs/2604.21375}, 
}

@misc{shinn2023reflexionlanguageagentsverbal,
      title={Reflexion: Language Agents with Verbal Reinforcement Learning}, 
      author={Noah Shinn and Federico Cassano and Edward Berman and Ashwin Gopinath and Karthik Narasimhan and Shunyu Yao},
      year={2023},
      eprint={2303.11366},
      archivePrefix={arXiv},
      primaryClass={cs.AI},
      url={https://arxiv.org/abs/2303.11366}, 
}

@misc{pan2024autonomousevaluationrefinementdigital,
      title={Autonomous Evaluation and Refinement of Digital Agents}, 
      author={Jiayi Pan and Yichi Zhang and Nicholas Tomlin and Yifei Zhou and Sergey Levine and Alane Suhr},
      year={2024},
      eprint={2404.06474},
      archivePrefix={arXiv},
      primaryClass={cs.AI},
      url={https://arxiv.org/abs/2404.06474}, 
}

@misc{wang2025opencuaopenfoundationscomputeruse,
      title={OpenCUA: Open Foundations for Computer-Use Agents}, 
      author={Xinyuan Wang and Bowen Wang and Dunjie Lu and Junlin Yang and Tianbao Xie and Junli Wang and Jiaqi Deng and Xiaole Guo and Yiheng Xu and Chen Henry Wu and Zhennan Shen and Zhuokai Li and Ryan Li and Xiaochuan Li and Junda Chen and Boyuan Zheng and Peihang Li and Fangyu Lei and Ruisheng Cao and Yeqiao Fu and Dongchan Shin and Martin Shin and Jiarui Hu and Yuyan Wang and Jixuan Chen and Yuxiao Ye and Danyang Zhang and Dikang Du and Hao Hu and Huarong Chen and Zaida Zhou and Haotian Yao and Ziwei Chen and Qizheng Gu and Yipu Wang and Heng Wang and Diyi Yang and Victor Zhong and Flood Sung and Y. Charles and Zhilin Yang and Tao Yu},
      year={2025},
      eprint={2508.09123},
      archivePrefix={arXiv},
      primaryClass={cs.AI},
      url={https://arxiv.org/abs/2508.09123}, 
}

@misc{qin2025uitarspioneeringautomatedgui,
      title={UI-TARS: Pioneering Automated GUI Interaction with Native Agents}, 
      author={Yujia Qin and Yining Ye and Junjie Fang and Haoming Wang and Shihao Liang and Shizuo Tian and Junda Zhang and Jiahao Li and Yunxin Li and Shijue Huang and Wanjun Zhong and Kuanye Li and Jiale Yang and Yu Miao and Woyu Lin and Longxiang Liu and Xu Jiang and Qianli Ma and Jingyu Li and Xiaojun Xiao and Kai Cai and Chuang Li and Yaowei Zheng and Chaolin Jin and Chen Li and Xiao Zhou and Minchao Wang and Haoli Chen and Zhaojian Li and Haihua Yang and Haifeng Liu and Feng Lin and Tao Peng and Xin Liu and Guang Shi},
      year={2025},
      eprint={2501.12326},
      archivePrefix={arXiv},
      primaryClass={cs.AI},
      url={https://arxiv.org/abs/2501.12326}, 
}

@misc{xue2026evocuaevolvingcomputeruse,
      title={EvoCUA: Evolving Computer Use Agents via Learning from Scalable Synthetic Experience}, 
      author={Taofeng Xue and Chong Peng and Mianqiu Huang and Linsen Guo and Tiancheng Han and Haozhe Wang and Jianing Wang and Xiaocheng Zhang and Xin Yang and Dengchang Zhao and Jinrui Ding and Xiandi Ma and Yuchen Xie and Peng Pei and Xunliang Cai and Xipeng Qiu},
      year={2026},
      eprint={2601.15876},
      archivePrefix={arXiv},
      primaryClass={cs.AI},
      url={https://arxiv.org/abs/2601.15876}, 
}

@misc{agashe2024agentsopenagentic,
      title={Agent S: An Open Agentic Framework that Uses Computers Like a Human}, 
      author={Saaket Agashe and Jiuzhou Han and Shuyu Gan and Jiachen Yang and Ang Li and Xin Eric Wang},
      year={2024},
      eprint={2410.08164},
      archivePrefix={arXiv},
      primaryClass={cs.AI},
      url={https://arxiv.org/abs/2410.08164}, 
}

@misc{agashe2025agents2compositionalgeneralistspecialist,
      title={Agent S2: A Compositional Generalist-Specialist Framework for Computer Use Agents}, 
      author={Saaket Agashe and Kyle Wong and Vincent Tu and Jiachen Yang and Ang Li and Xin Eric Wang},
      year={2025},
      eprint={2504.00906},
      archivePrefix={arXiv},
      primaryClass={cs.AI},
      url={https://arxiv.org/abs/2504.00906}, 
}

@misc{gonzalezpumariega2026scalingagentscomputeruse,
      title={Scaling Agents for Computer Use}, 
      author={Gonzalo Gonzalez-Pumariega and Vincent Tu and Chih-Lun Lee and Jiachen Yang and Ang Li and Xin Eric Wang},
      year={2026},
      eprint={2510.02250},
      archivePrefix={arXiv},
      primaryClass={cs.AI},
      url={https://arxiv.org/abs/2510.02250}, 
}

@misc{yang2025gta1guitesttimescaling,
      title={GTA1: GUI Test-time Scaling Agent}, 
      author={Yan Yang and Dongxu Li and Yutong Dai and Yuhao Yang and Ziyang Luo and Zirui Zhao and Zhiyuan Hu and Junzhe Huang and Amrita Saha and Zeyuan Chen and Ran Xu and Liyuan Pan and Silvio Savarese and Caiming Xiong and Junnan Li},
      year={2025},
      eprint={2507.05791},
      archivePrefix={arXiv},
      primaryClass={cs.AI},
      url={https://arxiv.org/abs/2507.05791}, 
}

@misc{xie2024osworldbenchmarkingmultimodalagents,
      title={OSWorld: Benchmarking Multimodal Agents for Open-Ended Tasks in Real Computer Environments}, 
      author={Tianbao Xie and Danyang Zhang and Jixuan Chen and Xiaochuan Li and Siheng Zhao and Ruisheng Cao and Toh Jing Hua and Zhoujun Cheng and Dongchan Shin and Fangyu Lei and Yitao Liu and Yiheng Xu and Shuyan Zhou and Silvio Savarese and Caiming Xiong and Victor Zhong and Tao Yu},
      year={2024},
      eprint={2404.07972},
      archivePrefix={arXiv},
      primaryClass={cs.AI},
      url={https://arxiv.org/abs/2404.07972}, 
}

@misc{liu2025paluiplanningactivelookback,
      title={PAL-UI: Planning with Active Look-back for Vision-Based GUI Agents}, 
      author={Zikang Liu and Junyi Li and Wayne Xin Zhao and Dawei Gao and Yaliang Li and Ji-rong Wen},
      year={2025},
      eprint={2510.00413},
      archivePrefix={arXiv},
      primaryClass={cs.CV},
      url={https://arxiv.org/abs/2510.00413}, 
}

@misc{yang2026ossymphonyholisticframeworkrobust,
      title={OS-Symphony: A Holistic Framework for Robust and Generalist Computer-Using Agent}, 
      author={Bowen Yang and Kaiming Jin and Zhenyu Wu and Zhaoyang Liu and Qiushi Sun and Zehao Li and JingJing Xie and Zhoumianze Liu and Fangzhi Xu and Kanzhi Cheng and Qingyun Li and Yian Wang and Yu Qiao and Zun Wang and Zichen Ding},
      year={2026},
      eprint={2601.07779},
      archivePrefix={arXiv},
      primaryClass={cs.MA},
      url={https://arxiv.org/abs/2601.07779}, 
}

@misc{song2026coact1computerusingmultiagentcoding,
      title={CoAct-1: Computer-using Multi-Agent System with Coding Actions}, 
      author={Linxin Song and Yutong Dai and Viraj Prabhu and Jieyu Zhang and Taiwei Shi and Li Li and Junnan Li and Silvio Savarese and Zeyuan Chen and Jieyu Zhao and Ran Xu and Caiming Xiong},
      year={2026},
      eprint={2508.03923},
      archivePrefix={arXiv},
      primaryClass={cs.CL},
      url={https://arxiv.org/abs/2508.03923}, 
}

@misc{opus4.6,
  author       = {{Anthropic}},
  title        = {Introducing Claude Opus 4.6},
  year         = {2026},
  howpublished = {\url{https://www.anthropic.com/news/claude-opus-4-6}},
}

@misc{sonnet4.6,
  author       = {{Anthropic}},
  title        = {Introducing Claude Sonnet 4.6},
  year         = {2026},
  howpublished = {\url{https://www.anthropic.com/news/claude-sonnet-4-6}},
}

@misc{gemini31pro,
  author       = {{Google}},
  title        = {Gemini 3.1 pro model card},
  year         = {2026},
  howpublished = {\url{https://storage.googleapis.com/deepmind-media/Model-Cards/Gemini-3-1-Pro-Model-Card.pdf}},
}

@misc{singh2026openaigpt5card,
      title={OpenAI GPT-5 System Card}, 
      author={Aaditya Singh and Adam Fry and Adam Perelman and Adam Tart and Adi Ganesh and Ahmed El-Kishky and Aidan McLaughlin and Aiden Low and AJ Ostrow and Akhila Ananthram and Akshay Nathan and Alan Luo and Alec Helyar and Aleksander Madry and Aleksandr Efremov and Aleksandra Spyra and Alex Baker-Whitcomb and Alex Beutel and Alex Karpenko and Alex Makelov and Alex Neitz and Alex Wei and Alexandra Barr and Alexandre Kirchmeyer and Alexey Ivanov and Alexi Christakis and Alistair Gillespie and Allison Tam and Ally Bennett and Alvin Wan and Alyssa Huang and Amy McDonald Sandjideh and Amy Yang and Ananya Kumar and Andre Saraiva and Andrea Vallone and Andrei Gheorghe and Andres Garcia Garcia and Andrew Braunstein and Andrew Liu and Andrew Schmidt and Andrey Mereskin and Andrey Mishchenko and Andy Applebaum and Andy Rogerson and Ann Rajan and Annie Wei and Anoop Kotha and Anubha Srivastava and Anushree Agrawal and Arun Vijayvergiya and Ashley Tyra and Ashvin Nair and Avi Nayak and Ben Eggers and Bessie Ji and Beth Hoover and Bill Chen and Blair Chen and Boaz Barak and Borys Minaiev and Botao Hao and Bowen Baker and Brad Lightcap and Brandon McKinzie and Brandon Wang and Brendan Quinn and Brian Fioca and Brian Hsu and Brian Yang and Brian Yu and Brian Zhang and Brittany Brenner and Callie Riggins Zetino and Cameron Raymond and Camillo Lugaresi and Carolina Paz and Cary Hudson and Cedric Whitney and Chak Li and Charles Chen and Charlotte Cole and Chelsea Voss and Chen Ding and Chen Shen and Chengdu Huang and Chris Colby and Chris Hallacy and Chris Koch and Chris Lu and Christina Kaplan and Christina Kim and CJ Minott-Henriques and Cliff Frey and Cody Yu and Coley Czarnecki and Colin Reid and Colin Wei and Cory Decareaux and Cristina Scheau and Cyril Zhang and Cyrus Forbes and Da Tang and Dakota Goldberg and Dan Roberts and Dana Palmie and Daniel Kappler and Daniel Levine and Daniel Wright and Dave Leo and David Lin and David Robinson and Declan Grabb and Derek Chen and Derek Lim and Derek Salama and Dibya Bhattacharjee and Dimitris Tsipras and Dinghua Li and Dingli Yu and DJ Strouse and Drew Williams and Dylan Hunn and Ed Bayes and Edwin Arbus and Ekin Akyurek and Elaine Ya Le and Elana Widmann and Eli Yani and Elizabeth Proehl and Enis Sert and Enoch Cheung and Eri Schwartz and Eric Han and Eric Jiang and Eric Mitchell and Eric Sigler and Eric Wallace and Erik Ritter and Erin Kavanaugh and Evan Mays and Evgenii Nikishin and Fangyuan Li and Felipe Petroski Such and Filipe de Avila Belbute Peres and Filippo Raso and Florent Bekerman and Foivos Tsimpourlas and Fotis Chantzis and Francis Song and Francis Zhang and Gaby Raila and Garrett McGrath and Gary Briggs and Gary Yang and Giambattista Parascandolo and Gildas Chabot and Grace Kim and Grace Zhao and Gregory Valiant and Guillaume Leclerc and Hadi Salman and Hanson Wang and Hao Sheng and Haoming Jiang and Haoyu Wang and Haozhun Jin and Harshit Sikchi and Heather Schmidt and Henry Aspegren and Honglin Chen and Huida Qiu and Hunter Lightman and Ian Covert and Ian Kivlichan and Ian Silber and Ian Sohl and Ibrahim Hammoud and Ignasi Clavera and Ikai Lan and Ilge Akkaya and Ilya Kostrikov and Irina Kofman and Isak Etinger and Ishaan Singal and Jackie Hehir and Jacob Huh and Jacqueline Pan and Jake Wilczynski and Jakub Pachocki and James Lee and James Quinn and Jamie Kiros and Janvi Kalra and Jasmyn Samaroo and Jason Wang and Jason Wolfe and Jay Chen and Jay Wang and Jean Harb and Jeffrey Han and Jeffrey Wang and Jennifer Zhao and Jeremy Chen and Jerene Yang and Jerry Tworek and Jesse Chand and Jessica Landon and Jessica Liang and Ji Lin and Jiancheng Liu and Jianfeng Wang and Jie Tang and Jihan Yin and Joanne Jang and Joel Morris and Joey Flynn and Johannes Ferstad and Johannes Heidecke and John Fishbein and John Hallman and Jonah Grant and Jonathan Chien and Jonathan Gordon and Jongsoo Park and Jordan Liss and Jos Kraaijeveld and Joseph Guay and Joseph Mo and Josh Lawson and Josh McGrath and Joshua Vendrow and Joy Jiao and Julian Lee and Julie Steele and Julie Wang and Junhua Mao and Kai Chen and Kai Hayashi and Kai Xiao and Kamyar Salahi and Kan Wu and Karan Sekhri and Karan Sharma and Karan Singhal and Karen Li and Kenny Nguyen and Keren Gu-Lemberg and Kevin King and Kevin Liu and Kevin Stone and Kevin Yu and Kristen Ying and Kristian Georgiev and Kristie Lim and Kushal Tirumala and Kyle Miller and Lama Ahmad and Larry Lv and Laura Clare and Laurance Fauconnet and Lauren Itow and Lauren Yang and Laurentia Romaniuk and Leah Anise and Lee Byron and Leher Pathak and Leon Maksin and Leyan Lo and Leyton Ho and Li Jing and Liang Wu and Liang Xiong and Lien Mamitsuka and Lin Yang and Lindsay McCallum and Lindsey Held and Liz Bourgeois and Logan Engstrom and Lorenz Kuhn and Louis Feuvrier and Lu Zhang and Lucas Switzer and Lukas Kondraciuk and Lukasz Kaiser and Manas Joglekar and Mandeep Singh and Mandip Shah and Manuka Stratta and Marcus Williams and Mark Chen and Mark Sun and Marselus Cayton and Martin Li and Marvin Zhang and Marwan Aljubeh and Matt Nichols and Matthew Haines and Max Schwarzer and Mayank Gupta and Meghan Shah and Melody Y. Guan and Melody Huang and Meng Dong and Mengqing Wang and Mia Glaese and Micah Carroll and Michael Lampe and Michael Malek and Michael Sharman and Michael Zhang and Michele Wang and Michelle Pokrass and Mihai Florian and Mikhail Pavlov and Miles Wang and Ming Chen and Mingxuan Wang and Minnia Feng and Mo Bavarian and Molly Lin and Moose Abdool and Mostafa Rohaninejad and Nacho Soto and Natalie Staudacher and Natan LaFontaine and Nathan Marwell and Nelson Liu and Nick Preston and Nick Turley and Nicklas Ansman and Nicole Blades and Nikil Pancha and Nikita Mikhaylin and Niko Felix and Nikunj Handa and Nishant Rai and Nitish Keskar and Noam Brown and Ofir Nachum and Oleg Boiko and Oleg Murk and Olivia Watkins and Oona Gleeson and Pamela Mishkin and Patryk Lesiewicz and Paul Baltescu and Pavel Belov and Peter Zhokhov and Philip Pronin and Phillip Guo and Phoebe Thacker and Qi Liu and Qiming Yuan and Qinghua Liu and Rachel Dias and Rachel Puckett and Rahul Arora and Ravi Teja Mullapudi and Raz Gaon and Reah Miyara and Rennie Song and Rishabh Aggarwal and RJ Marsan and Robel Yemiru and Robert Xiong and Rohan Kshirsagar and Rohan Nuttall and Roman Tsiupa and Ronen Eldan and Rose Wang and Roshan James and Roy Ziv and Rui Shu and Ruslan Nigmatullin and Saachi Jain and Saam Talaie and Sam Altman and Sam Arnesen and Sam Toizer and Sam Toyer and Samuel Miserendino and Sandhini Agarwal and Sarah Yoo and Savannah Heon and Scott Ethersmith and Sean Grove and Sean Taylor and Sebastien Bubeck and Sever Banesiu and Shaokyi Amdo and Shengjia Zhao and Sherwin Wu and Shibani Santurkar and Shiyu Zhao and Shraman Ray Chaudhuri and Shreyas Krishnaswamy and Shuaiqi and Xia and Shuyang Cheng and Shyamal Anadkat and Simón Posada Fishman and Simon Tobin and Siyuan Fu and Somay Jain and Song Mei and Sonya Egoian and Spencer Kim and Spug Golden and SQ Mah and Steph Lin and Stephen Imm and Steve Sharpe and Steve Yadlowsky and Sulman Choudhry and Sungwon Eum and Suvansh Sanjeev and Tabarak Khan and Tal Stramer and Tao Wang and Tao Xin and Tarun Gogineni and Taya Christianson and Ted Sanders and Tejal Patwardhan and Thomas Degry and Thomas Shadwell and Tianfu Fu and Tianshi Gao and Timur Garipov and Tina Sriskandarajah and Toki Sherbakov and Tomek Korbak and Tomer Kaftan and Tomo Hiratsuka and Tongzhou Wang and Tony Song and Tony Zhao and Troy Peterson and Val Kharitonov and Victoria Chernova and Vineet Kosaraju and Vishal Kuo and Vitchyr Pong and Vivek Verma and Vlad Petrov and Wanning Jiang and Weixing Zhang and Wenda Zhou and Wenlei Xie and Wenting Zhan and Wes McCabe and Will DePue and Will Ellsworth and Wulfie Bain and Wyatt Thompson and Xiangning Chen and Xiangyu Qi and Xin Xiang and Xinwei Shi and Yann Dubois and Yaodong Yu and Yara Khakbaz and Yifan Wu and Yilei Qian and Yin Tat Lee and Yinbo Chen and Yizhen Zhang and Yizhong Xiong and Yonglong Tian and Young Cha and Yu Bai and Yu Yang and Yuan Yuan and Yuanzhi Li and Yufeng Zhang and Yuguang Yang and Yujia Jin and Yun Jiang and Yunyun Wang and Yushi Wang and Yutian Liu and Zach Stubenvoll and Zehao Dou and Zheng Wu and Zhigang Wang},
      year={2026},
      eprint={2601.03267},
      archivePrefix={arXiv},
      primaryClass={cs.CL},
      url={https://arxiv.org/abs/2601.03267}, 
}

@misc{bai2025qwen3vltechnicalreport,
      title={Qwen3-VL Technical Report}, 
      author={Shuai Bai and Yuxuan Cai and Ruizhe Chen and Keqin Chen and Xionghui Chen and Zesen Cheng and Lianghao Deng and Wei Ding and Chang Gao and Chunjiang Ge and Wenbin Ge and Zhifang Guo and Qidong Huang and Jie Huang and Fei Huang and Binyuan Hui and Shutong Jiang and Zhaohai Li and Mingsheng Li and Mei Li and Kaixin Li and Zicheng Lin and Junyang Lin and Xuejing Liu and Jiawei Liu and Chenglong Liu and Yang Liu and Dayiheng Liu and Shixuan Liu and Dunjie Lu and Ruilin Luo and Chenxu Lv and Rui Men and Lingchen Meng and Xuancheng Ren and Xingzhang Ren and Sibo Song and Yuchong Sun and Jun Tang and Jianhong Tu and Jianqiang Wan and Peng Wang and Pengfei Wang and Qiuyue Wang and Yuxuan Wang and Tianbao Xie and Yiheng Xu and Haiyang Xu and Jin Xu and Zhibo Yang and Mingkun Yang and Jianxin Yang and An Yang and Bowen Yu and Fei Zhang and Hang Zhang and Xi Zhang and Bo Zheng and Humen Zhong and Jingren Zhou and Fan Zhou and Jing Zhou and Yuanzhi Zhu and Ke Zhu},
      year={2025},
      eprint={2511.21631},
      archivePrefix={arXiv},
      primaryClass={cs.CV},
      url={https://arxiv.org/abs/2511.21631}, 
}

@misc{xu2025aguvisunifiedpurevision,
      title={Aguvis: Unified Pure Vision Agents for Autonomous GUI Interaction}, 
      author={Yiheng Xu and Zekun Wang and Junli Wang and Dunjie Lu and Tianbao Xie and Amrita Saha and Doyen Sahoo and Tao Yu and Caiming Xiong},
      year={2025},
      eprint={2412.04454},
      archivePrefix={arXiv},
      primaryClass={cs.CL},
      url={https://arxiv.org/abs/2412.04454}, 
}

@misc{lin2024showuivisionlanguageactionmodelgui,
      title={ShowUI: One Vision-Language-Action Model for GUI Visual Agent}, 
      author={Kevin Qinghong Lin and Linjie Li and Difei Gao and Zhengyuan Yang and Shiwei Wu and Zechen Bai and Weixian Lei and Lijuan Wang and Mike Zheng Shou},
      year={2024},
      eprint={2411.17465},
      archivePrefix={arXiv},
      primaryClass={cs.CV},
      url={https://arxiv.org/abs/2411.17465}, 
}

@misc{deng2023mind2webgeneralistagentweb,
      title={Mind2Web: Towards a Generalist Agent for the Web}, 
      author={Xiang Deng and Yu Gu and Boyuan Zheng and Shijie Chen and Samuel Stevens and Boshi Wang and Huan Sun and Yu Su},
      year={2023},
      eprint={2306.06070},
      archivePrefix={arXiv},
      primaryClass={cs.CL},
      url={https://arxiv.org/abs/2306.06070}, 
}

@misc{zhou2024webarenarealisticwebenvironment,
      title={WebArena: A Realistic Web Environment for Building Autonomous Agents}, 
      author={Shuyan Zhou and Frank F. Xu and Hao Zhu and Xuhui Zhou and Robert Lo and Abishek Sridhar and Xianyi Cheng and Tianyue Ou and Yonatan Bisk and Daniel Fried and Uri Alon and Graham Neubig},
      year={2024},
      eprint={2307.13854},
      archivePrefix={arXiv},
      primaryClass={cs.AI},
      url={https://arxiv.org/abs/2307.13854}, 
}

@misc{bonatti2024windowsagentarenaevaluating,
      title={Windows Agent Arena: Evaluating Multi-Modal OS Agents at Scale}, 
      author={Rogerio Bonatti and Dan Zhao and Francesco Bonacci and Dillon Dupont and Sara Abdali and Yinheng Li and Yadong Lu and Justin Wagle and Kazuhito Koishida and Arthur Bucker and Lawrence Jang and Zack Hui},
      year={2024},
      eprint={2409.08264},
      archivePrefix={arXiv},
      primaryClass={cs.AI},
      url={https://arxiv.org/abs/2409.08264}, 
}

@misc{rawles2025androidworlddynamicbenchmarkingenvironment,
      title={AndroidWorld: A Dynamic Benchmarking Environment for Autonomous Agents}, 
      author={Christopher Rawles and Sarah Clinckemaillie and Yifan Chang and Jonathan Waltz and Gabrielle Lau and Marybeth Fair and Alice Li and William Bishop and Wei Li and Folawiyo Campbell-Ajala and Daniel Toyama and Robert Berry and Divya Tyamagundlu and Timothy Lillicrap and Oriana Riva},
      year={2025},
      eprint={2405.14573},
      archivePrefix={arXiv},
      primaryClass={cs.AI},
      url={https://arxiv.org/abs/2405.14573}, 
}

@misc{wang2023selfconsistencyimproveschainthought,
      title={Self-Consistency Improves Chain of Thought Reasoning in Language Models}, 
      author={Xuezhi Wang and Jason Wei and Dale Schuurmans and Quoc Le and Ed Chi and Sharan Narang and Aakanksha Chowdhery and Denny Zhou},
      year={2023},
      eprint={2203.11171},
      archivePrefix={arXiv},
      primaryClass={cs.CL},
      url={https://arxiv.org/abs/2203.11171}, 
}

@misc{yao2023treethoughtsdeliberateproblem,
      title={Tree of Thoughts: Deliberate Problem Solving with Large Language Models}, 
      author={Shunyu Yao and Dian Yu and Jeffrey Zhao and Izhak Shafran and Thomas L. Griffiths and Yuan Cao and Karthik Narasimhan},
      year={2023},
      eprint={2305.10601},
      archivePrefix={arXiv},
      primaryClass={cs.CL},
      url={https://arxiv.org/abs/2305.10601}, 
}

@misc{lin2026uivoyagerselfevolvingguiagent,
      title={UI-Voyager: A Self-Evolving GUI Agent Learning via Failed Experience}, 
      author={Zichuan Lin and Feiyu Liu and Yijun Yang and Jiafei Lyu and Yiming Gao and Yicheng Liu and Zhicong Lu and Yangbin Yu and Mingyu Yang and Junyou Li and Deheng Ye and Jie Jiang},
      year={2026},
      eprint={2603.24533},
      archivePrefix={arXiv},
      primaryClass={cs.LG},
      url={https://arxiv.org/abs/2603.24533}, 
}

@misc{madaan2023selfrefineiterativerefinementselffeedback,
      title={Self-Refine: Iterative Refinement with Self-Feedback}, 
      author={Aman Madaan and Niket Tandon and Prakhar Gupta and Skyler Hallinan and Luyu Gao and Sarah Wiegreffe and Uri Alon and Nouha Dziri and Shrimai Prabhumoye and Yiming Yang and Shashank Gupta and Bodhisattwa Prasad Majumder and Katherine Hermann and Sean Welleck and Amir Yazdanbakhsh and Peter Clark},
      year={2023},
      eprint={2303.17651},
      archivePrefix={arXiv},
      primaryClass={cs.CL},
      url={https://arxiv.org/abs/2303.17651}, 
}

@misc{zhu2025scalingtesttimecomputellm,
      title={Scaling Test-time Compute for LLM Agents}, 
      author={King Zhu and Hanhao Li and Siwei Wu and Tianshun Xing and Dehua Ma and Xiangru Tang and Minghao Liu and Jian Yang and Jiaheng Liu and Yuchen Eleanor Jiang and Changwang Zhang and Chenghua Lin and Jun Wang and Ge Zhang and Wangchunshu Zhou},
      year={2025},
      eprint={2506.12928},
      archivePrefix={arXiv},
      primaryClass={cs.AI},
      url={https://arxiv.org/abs/2506.12928}, 
}

@misc{lee2026agenticaggregationparallelscaling,
      title={Agentic Aggregation for Parallel Scaling of Long-Horizon Agentic Tasks}, 
      author={Yoonsang Lee and Howard Yen and Xi Ye and Danqi Chen},
      year={2026},
      eprint={2604.11753},
      archivePrefix={arXiv},
      primaryClass={cs.CL},
      url={https://arxiv.org/abs/2604.11753}, 
}

@misc{belcak2025smalllanguagemodelsfuture,
      title={Small Language Models are the Future of Agentic AI}, 
      author={Peter Belcak and Greg Heinrich and Shizhe Diao and Yonggan Fu and Xin Dong and Saurav Muralidharan and Yingyan Celine Lin and Pavlo Molchanov},
      year={2025},
      eprint={2506.02153},
      archivePrefix={arXiv},
      primaryClass={cs.AI},
      url={https://arxiv.org/abs/2506.02153}, 
}

\clearpage

\appendix

\section{CANDID Implementation}

\subsection{Loop Detection Details}
\label{sec:appendix-loop-detection}

This section details the loop detection procedure used by the loop-recovery module in Section~\ref{sec:method-loop}. The implementation follows a two-stage detection. First, a deterministic rule detector checks whether the newly proposed action would extend a recent repetitive pattern. Second, the local model confirms that the trajectory is stagnant before recovery candidates are generated.

\paragraph{Action signatures.}
For each proposed action, the detector constructs a coarse signature consisting of an action type, an optional target coordinate, and a normalized action string. Action types are extracted from parsed PyAutoGUI calls when available: click-like calls are mapped to click variants, \texttt{typewrite}/\texttt{write} to \texttt{type}, \texttt{hotkey}/\texttt{press} to \texttt{key}, \texttt{scroll}/\texttt{hscroll} to \texttt{scroll}, \texttt{moveTo} to \texttt{mouse\_move}, and \texttt{dragTo} to \texttt{drag}. If the parsed code is unavailable, the detector falls back to coarse markers in the natural language action string. Target coordinates are extracted from the first \texttt{(x, y)} pair when present.

\begin{table*}[t]
\centering
\small
\begin{tabular}
{p{0.35\linewidth}p{0.35\linewidth}p{0.2\linewidth}}
\toprule
\textbf{Trigger reason} & \textbf{Condition} & \textbf{Intuition} \\
\midrule
\texttt{repeated\_action\_type\_and\_target}
& The proposed action has a target coordinate, and at least $r-1$ recent actions
within the window have the same action type and a target within tolerance
$\tau$ of the proposed target.
& The agent is repeatedly operating on the same local screen region. \\
\texttt{repeated\_targetless\_action\_type}
& The proposed action type is \texttt{scroll} or \texttt{wait}, and at least
$r-1$ recent actions in the window have the same action type.
& The agent is repeatedly scrolling or waiting without a concrete target. \\
\texttt{repeated\_exact\_text\_or\_key\_action}
& The proposed action type is \texttt{type} or \texttt{key}, and the normalized
full action string appears at least $r-1$ times in the recent window.
& Text entry and keyboard commands require exact matching because the payload
matters. \\
\texttt{repeated\_response\_and\_action\_type}
& The previous two assistant responses are identical after normalization, and
there is at least one recent action with the same action type as the proposed
action.
& The model is reproducing the same reasoning or action description while
continuing the same type of operation. \\
\texttt{repeated\_cyclic\_action\_pattern}
& The recent action-signature sequence plus the proposed action ends with a
repeated pattern of length 2--4.
& The agent alternates among a small set of actions rather than repeating one
literal action. \\
\bottomrule
\end{tabular}
\caption{Rule-based loop trigger conditions. Here $r$ is
\texttt{loop\_recovery\_min\_repeats}, $\tau$ is
\texttt{loop\_recovery\_target\_tolerance}, and the recent action window is
\texttt{loop\_recovery\_window}. In our main results, these are
$r=3$, $\tau=80$, and a window of 5 actions.}
\label{tab:loop-detection-rules}
\end{table*}

\paragraph{Rule-based trigger.}
Let $a_t$ be the action proposed for the current step. If the parsed action is empty or is a termination action, loop recovery is not triggered. Otherwise, the detector compares $a_t$ against the last $w$ executed actions, where $w=\texttt{loop\_recovery\_window}$. The detector requires at least $r-1$ historical actions before testing repetition, because the current proposal itself supplies the $r$-th repeated item. The concrete trigger conditions are listed in Table~\ref{tab:loop-detection-rules}.

For cyclic patterns, the detector uses a slightly longer window, $\max(w,8)$, appends the current proposal, and converts each action into a compact sequence item. Coordinate-based actions are bucketed by the target tolerance $\tau$; \texttt{type} and \texttt{key} actions retain their normalized full action string; all other actions use only the coarse action type. The detector then checks whether the sequence suffix repeats with a pattern length from 2 to 4. This catches cases such as alternating between two nearby clicks or repeatedly cycling through scroll and click operations.

\paragraph{Model confirmation gate.}
The rule detector is intentionally high-recall: it identifies points where repetition is plausible, but it does not independently decide that the trajectory is unproductive. A model confirmation call therefore follows a positive rule trigger. The confirmation prompt contains the task, current screenshot, recent trajectory, and the current proposed action, and asks the same local model to return a JSON object with progress\_signal, loop\_detected, loop\_evidence, strategy\_signal, avoid\_actions, and escape\_direction. Recovery proceeds only when either \texttt{loop\_detected=true} or \texttt{strategy\_signal=SWITCH}. Otherwise, the original action is left unchanged and candidate generation is skipped.

\paragraph{Candidate blocking after detection.}
The detected rule also defines what recovery candidates are forbidden to repeat. Malformed candidates are removed, and \texttt{DONE} termination is disallowed during loop recovery. If a candidate has a different action type from the blocked action type, it remains eligible. If it has the same action type, it is blocked when it lacks a distinguishable target or when its target lies within the same tolerance $\tau$ of the blocked target. This ensures that recovery does not simply reselect the same local operation under a different surface form.

\subsection{CANDID Prompts}
\label{sec:appendix-candid-prompts}

Below we provide the prompt skeletons used by CANDID. Runtime fields are shown in angle brackets. The normal action-generation prompt includes the computer-use tool schema and requires one \texttt{Action} line followed by one \texttt{\textless tool\_call\textgreater} block.

\subsubsection{Loop-Recovery Candidate Selection}

\paragraph{Loop-confirmation prompt.}
After the rule-based detector flags a potential repetitive loop, CANDID asks the local model to independently confirm that the proposed action would continue a no-progress loop. Rule-detector evidence is deliberately omitted from this prompt to avoid biasing the confirmation.
\begin{center}
\setlength{\fboxsep}{6pt}
\fbox{%
\begin{minipage}{0.92\linewidth}
\small\raggedright
\textbf{System Prompt}\par
\smallskip
You are a loop-confirmation agent for a desktop automation task. Analyze the recent trajectory and current screenshot for repetitive no-progress behavior. Focus only on loop recovery. Do not judge whether the whole task is complete. Return only a valid JSON object.

\medskip
\textbf{User Prompt}\par
\smallskip
\textbf{Task description:} \texttt{<TASK INSTRUCTION>}\par
\smallskip
\textbf{Recent trajectory before the current proposed action:}\\
\texttt{<RECENT TRAJECTORY>}\par
\smallskip
\textbf{Current proposed action before execution:}\\
\textbf{Assistant response:} \texttt{<MODEL RESPONSE>}\\
\textbf{Parsed action:} \texttt{<PARSED ACTION>}\\
\textbf{PyAutoGUI code:} \texttt{<PYAUTOGUI CODE>}\par
\smallskip
Assess whether executing the current proposed action would continue a repetitive no-progress loop. Focus only on loop recovery. Do not judge whether the whole task is complete. Return JSON exactly with keys:
\begin{quote}
\ttfamily
\{"progress\_signal": "what visibly changed, or no visible progress",\\
"loop\_detected": true/false,\\
"loop\_evidence": "short evidence",\\
"strategy\_signal": "KEEP or SWITCH",\\
"avoid\_actions": ["short action/target patterns to avoid"],\\
"escape\_direction": "short direction for changing approach"\}
\end{quote}
\textit{Multimodal input: current screenshot.}
\end{minipage}}
\end{center}
Recovery proceeds only if the returned \texttt{loop\_detected} value is true or the \texttt{strategy\_signal} is \texttt{SWITCH}.

\paragraph{Loop-recovery candidate-generation prompt.}
For each recovery sample, CANDID reuses the normal action-generation system
prompt, tool schema, recent multimodal history, and current screenshot. It
appends the following control block to the current user message:
\begin{center}
\setlength{\fboxsep}{6pt}
\fbox{%
\begin{minipage}{0.92\linewidth}
\small\raggedright
\textbf{Control Block Appended to the User Prompt}\par
\smallskip
\textbf{Loop recovery note:}\\
The recent trajectory status: \texttt{<PROGRESS SIGNAL>}

\smallskip
\textbf{Avoid:}\\
\textbullet~\texttt{<AVOID ACTIONS>}

\smallskip
\textbf{Try instead:}\\
\textbullet~choose a different target, menu path, keyboard shortcut, or
navigation strategy\\
\textbullet~\texttt{<ESCAPE DIRECTION>}

\smallskip
\textbf{Evidence:}\\
\texttt{<LOOP EVIDENCE>}

\smallskip
Generate the next action using the normal action format. Do not terminate or
answer \texttt{DONE} during loop recovery.
\end{minipage}}
\end{center}
Each sampled output retains the base action format:
\begin{center}
\setlength{\fboxsep}{6pt}
\fbox{%
\begin{minipage}{0.92\linewidth}
\small\ttfamily\raggedright
Action: <SHORT IMPERATIVE ACTION>\\
\textless tool\_call\textgreater\\
\{"name": "computer\_use", "arguments": \{<ACTION ARGUMENTS>\}\}\\
\textless/tool\_call\textgreater
\end{minipage}}
\end{center}

\paragraph{Loop-recovery candidate-selection prompt.}
Malformed, duplicate, and blocked candidates are removed before selection. The
selector receives only the remaining candidates.
\begin{center}
\setlength{\fboxsep}{6pt}
\fbox{%
\begin{minipage}{0.92\linewidth}
\small\raggedright
\textbf{System Prompt}\par
\smallskip
You select the best loop-recovery action for a desktop automation agent. The normal action model appears stuck in a repeated local pattern. Choose exactly one provided candidate. Do not invent a new action. Prefer a candidate that changes target, action type, menu path, or modality while still making task progress. Do not choose an action just because it is different; it must be useful for the current screen. Return only a valid JSON object.

\medskip
\textbf{User Prompt}\par
\smallskip
\textit{Multimodal input: current screenshot.}\par
\smallskip
\textbf{Task instruction:} \texttt{<TASK INSTRUCTION>}\par
\textbf{Recent trajectory:} \texttt{<RECENT TRAJECTORY>}\par
\smallskip
\textbf{Loop evidence:}\\
\textbullet~reason: \texttt{<TRIGGER REASON>}\\
\textbullet~proposed repeated action: \texttt{<PROPOSED ACTION>}\\
\textbullet~blocked action type: \texttt{<BLOCKED ACTION TYPE>}\\
\textbullet~blocked target: \texttt{<BLOCKED TARGET>}\par
\smallskip
\textbf{Candidate recovery actions:}\par
\textbf{Candidate \texttt{<ID>}:}\\
Parsed action: \texttt{<PARSED ACTION>}\\
PyAutoGUI code: \texttt{<PYAUTOGUI CODE>}\\
Action type: \texttt{<ACTION TYPE>}\\
Target: \texttt{<TARGET>}\\
Model response: \texttt{<MODEL RESPONSE>}\par
\smallskip
\texttt{<ADDITIONAL CANDIDATES>}\par
\smallskip
Return JSON exactly with keys:
\begin{quote}
\ttfamily
\{"selected\_candidate\_id": <integer>,\\
"selected\_action": "candidate parsed action",\\
"reason": "short reason"\}
\end{quote}
\end{minipage}}
\end{center}

\subsubsection{Unanimous \texttt{DONE} Gating}

\paragraph{Candidate-generation prompt.}
When the initial action is \texttt{DONE}, additional candidates are sampled
from the base action-generation prompt. Thus, there is no separate
DONE-specific generation instruction. The prompt contains the computer-use
system prompt and tool schema, recent screenshot--response history, the current
screenshot, and the following current-step instruction:
\begin{center}
\setlength{\fboxsep}{6pt}
\fbox{%
\begin{minipage}{0.92\linewidth}
\small\raggedright
\textbf{Current-Step User Prompt}\par
\smallskip
Please generate the next move according to the UI screenshot, instruction and
previous actions.

\smallskip
\textbf{Instruction:} \texttt{<TASK INSTRUCTION>}

\smallskip
\textbf{Previous actions:}\\
\texttt{<PREVIOUS ACTIONS>}

\smallskip
\textit{Multimodal input: current screenshot.}
\end{minipage}}
\end{center}

\paragraph{Candidate-selection prompt.}
The original \texttt{DONE} proposal and the additional samples are first
deduplicated by executable action. Under the unanimous policy, if any valid
non-\texttt{DONE} sample exists, all \texttt{DONE} candidates are excluded.
The following selector prompt is used only when more than one candidate remains
after this filtering:
\begin{center}
\setlength{\fboxsep}{6pt}
\fbox{%
\begin{minipage}{0.92\linewidth}
\small\raggedright
\textbf{System Prompt}\par
\smallskip
You select the best next action for a desktop automation agent. Compare the
candidates against the task instruction, current screenshot, and recent
trajectory. Choose exactly one provided candidate. Do not invent a new action.
Choose the candidate that is most correct for the current state and most likely
to complete the task successfully. If a candidate claims the task is complete,
select it only when the visible state and trajectory clearly support that
claim. Return only a valid JSON object.

\medskip
\textbf{User Prompt}\par
\smallskip
\textit{Multimodal input: current screenshot.}\par
\smallskip
\textbf{Task instruction:} \texttt{<TASK INSTRUCTION>}\par
\textbf{Recent trajectory:} \texttt{<RECENT TRAJECTORY>}\par
\smallskip
\textbf{Candidate actions:}\par
\textbf{Candidate \texttt{<ID>}:}\\
Parsed action: \texttt{<PARSED ACTION>}\\
PyAutoGUI code: \texttt{<PYAUTOGUI CODE>}\\
Is \texttt{DONE}: \texttt{<TRUE OR FALSE>}\\
Model response: \texttt{<MODEL RESPONSE>}\par
\smallskip
\texttt{<ADDITIONAL CANDIDATES>}\par
\smallskip
Return JSON exactly with keys:
\begin{quote}
\ttfamily
\{"selected\_candidate\_id": <integer>,\\
"selected\_action": "candidate parsed action",\\
"reason": "short reason"\}
\end{quote}
\end{minipage}}
\end{center}
If filtering leaves a single valid non-\texttt{DONE} candidate, it is selected
directly without an additional model call. If all samples agree on
\texttt{DONE}, termination is retained.

\subsection{Computing Infrastructure}
All experiments were conducted on a Linux (Ubuntu) server with an NVIDIA A100 80GB GPU. Model serving was performed using vLLM (v0.12.0) on top of the Transformers library (v5.2.0).

\section{Additional Analysis}

\subsection{Candidate Diversity}
\label{sec:candidate-diversity}
We analyze candidate diversity at triggered control points by counting the number of unique exact actions and unique action types before filtering. As shown in Table~\ref{tab:candidate-diversity}, Loop Recovery exhibits substantial diversity at the exact-action level: among 2,496 triggered events, all four sampled candidates are distinct in 1,190 cases (47.7\%), while only 209 cases (8.4\%) produce identical actions. In contrast, action-type diversity is much lower, with 2,095 events (83.9\%) sharing a single action type across all candidates. Similarly, DONE Gating frequently generates multiple distinct exact actions, with only 238 of 429 triggered events (55.5\%) producing identical candidates. These results suggest that candidate diversity primarily arises from differences in action parameters, such as click locations or typed text, rather than coarse GUI action types.

\begin{table}[t]
\centering
\small
\resizebox{\linewidth}{!}{%
\begin{tabular}{llrrrr}
\toprule
Module & Diversity Metric & 1 & 2 & 3 & 4 \\
\midrule
Loop Recovery & Exact action & 209 & 435 & 662 & 1,190 \\
Loop Recovery & Action type & 2,095 & 389 & 12 & 0 \\
DONE Gating & Exact action & 238 & 96 & 73 & 22 \\
\bottomrule
\end{tabular}
}
\caption{Candidate diversity at triggered control points.}
\label{tab:candidate-diversity}
\end{table}

\subsection{Loop Detector Hyperparameter Sensitivity}
\label{sec:loop-param}

\begin{table}[t]
\centering
\begin{tabular}{ccccc}
\toprule
$w$ & $\tau$ & $r$ & Trials & Overall (\%) \\
\midrule
5 & 80  & 3 & 5 & 34.84 $\pm$ 1.27 \\
3 & 80  & 3 & 3 & 36.05 $\pm$ 0.71 \\
7 & 80  & 3 & 3 & 35.94 $\pm$ 0.85 \\
5 & 0   & 3 & 3 & 36.88 $\pm$ 0.44 \\
5 & 40  & 3 & 3 & 35.02 $\pm$ 1.28 \\
5 & 120 & 3 & 3 & 34.96 $\pm$ 0.67 \\
5 & 80  & 2 & 3 & 33.04 $\pm$ 1.03 \\
5 & 80  & 4 & 3 & 34.85 $\pm$ 0.53 \\
5 & 80  & 5 & 3 & 35.01 $\pm$ 1.24 \\
\bottomrule
\end{tabular}
\caption{Sensitivity analysis of the loop detector hyperparameters using Qwen3-VL-8B with a step budget of $s=50$. Results are reported as mean task accuracy (\%) $\pm$ standard deviation over multiple trials.}
\label{tab:loop_detector_sensitivity}
\end{table}

Table~\ref{tab:loop_detector_sensitivity} varies the loop-detection window ($w$), target tolerance ($\tau$), and minimum repeat count ($r$). Although several three-trial settings yield numerically higher means than the five-trial main result, these gains are not consistent across neighboring values and may reflect stochastic variation in agent execution. In contrast, setting $r=2$ reduces accuracy to 33.04\%, nearly eliminating the gain over the 32.82\% baseline, suggesting that overly sensitive loop detection can interrupt legitimate repeated actions. Overall, performance remains broadly stable across moderate changes in the window and tolerance.

\subsection{Task-Level Analysis of Control Decisions}
\label{sec:appendix-control-analysis}

This section provides a task-level analysis of how CANDID's control modules behave after being triggered, with particular attention to whether interventions alter the original decision and how effectively the DONE gate distinguishes successful from premature termination.

OSWorld does not provide ground-truth labels indicating whether an individual recovery or termination decision is correct at each execution step. We therefore analyze control decisions using final task success as a task-level proxy. These results should not be interpreted as action-level correctness; rather, they characterize whether trajectories eventually succeed after a control module is triggered.

\subsubsection{Intervention Outcomes}

We first examine whether each triggered control decision changes the agent's original action. For Loop Recovery, a positive decision indicates that the original action is replaced by a recovery action. For Unanimous DONE Gating, a positive decision indicates that the original \texttt{DONE} proposal is redirected to a non-\texttt{DONE} continuation. We define TP as a changed decision followed by task success, FP as a changed decision followed by task failure, TN as an unchanged decision followed by task success, and FN as an unchanged decision followed by task failure. Table~\ref{tab:control-confusion} summarizes the resulting task-level outcomes for both control modules.

\begin{table}[t]
\centering
\small
\begin{tabular}{lrrrr}
\toprule
Method & TP & FP & TN & FN \\
\midrule
Loop Recovery & 26 & 148 & 8 & 10 \\
Unanimous DONE Gating & 18 & 67 & 114 & 70 \\
\bottomrule
\end{tabular}
\caption{Task-level outcomes of triggered control decisions. Final task success is used as a proxy because OSWorld does not provide per-step ground-truth control labels.}
\label{tab:control-confusion}
\end{table}

Among the 34 successful tasks in which Loop Recovery was triggered, 26 tasks (76.5\%) contained at least one action-changing intervention. This suggests that recovery actions are frequently involved in successful triggered trajectories. The relatively large FP count should be interpreted cautiously: Loop Recovery is activated primarily on difficult trajectories that already exhibit repetitive or stagnant behavior, and such tasks may fail eventually even when the local recovery action is reasonable.

For Unanimous DONE Gating, the large TN count indicates that the module frequently preserves the original termination decision in trajectories that ultimately succeed. This behavior is consistent with its role as a conservative policy: it redirects termination only when candidate disagreement provides evidence for continued execution.

\subsubsection{Termination-Gate Evaluation}
\label{sec:termination-gate}
We further compare Unanimous DONE Gating with the VLAA-GUI-style Verifier as termination gates. In this analysis, TP denotes an accepted \texttt{DONE} decision followed by task success, FP denotes an accepted \texttt{DONE} decision followed by task failure, TN denotes a rejected \texttt{DONE} decision followed by task failure, and FN denotes a rejected \texttt{DONE} decision followed by task success. Table~\ref{tab:done-gate-counts} reports the corresponding decision counts, and Table~\ref{tab:done-gate-metrics} presents the derived precision, recall, specificity, and accuracy.

\begin{table}[t]
\centering
\resizebox{\linewidth}{!}{%
\begin{tabular}{lrrrrr}
\toprule
Method & Triggered & TP & FP & TN & FN \\
\midrule
VLAA-GUI Verifier & 229 & 113 & 95 & 17 & 4 \\
Unanimous DONE Gating & 269 & 132 & 101 & 36 & 0 \\
\bottomrule
\end{tabular}
}
\caption{Task-level comparison of termination-gate decisions on Qwen3-VL-30B-A3B with a 50-step budget.}
\label{tab:done-gate-counts}
\end{table}

\begin{table}[t]
\centering
\small
\resizebox{\linewidth}{!}{%
\begin{tabular}{lrrrr}
\toprule
Method & Precision & Recall & Specificity & Accuracy \\
\midrule
VLAA-GUI Verifier
& 54.3\% & 96.6\% & 15.2\% & 56.8\% \\
Unanimous DONE Gating
& 56.7\% & 100.0\% & 26.3\% & 62.5\% \\
\bottomrule
\end{tabular}
}
\caption{Precision, recall, specificity, and accuracy of the termination gates.}
\label{tab:done-gate-metrics}
\end{table}

Both methods achieve high recall, indicating that tasks that ultimately succeed are rarely prevented from terminating. However, precision remains modest for both gates, meaning that a substantial fraction of accepted \texttt{DONE} decisions still occur in trajectories that fail according to the OSWorld evaluator. This limitation reflects the difficulty of inferring semantic task completion from the local model's own observations and predictions.

The main difference appears in specificity. Unanimous DONE Gating achieves higher specificity than the VLAA-GUI Verifier (26.3\% vs.\ 15.2\%), indicating that it rejects a larger fraction of premature termination proposals. This improvement also leads to higher task-level decision accuracy (62.5\% vs.\ 56.8\%). Overall, these results suggest that the unanimous policy provides a more effective sampling-based check against unsupported termination than textual verification.

\begin{table*}[t]
\centering
\small
\resizebox{\textwidth}{!}{%
\begin{tabular}{llccccccc}
\toprule
\multirow{2}{*}{\textbf{Model}} & \multirow{2}{*}{\textbf{Method}} & \multicolumn{6}{c}{Domain Accuracy (\%)} & \multirow{2}{*}{\textbf{Overall (\%)}} \\
\cmidrule(lr){3-8}
& & \textbf{Office} (43) & \textbf{Web} (30) & \textbf{Windows System} (24) & \textbf{Coding} (24) & \textbf{Media} (21) & \textbf{Windows Utilities} (12) & \\
\midrule
\multirow{2}{*}{Qwen3-VL-30B-A3B}
& Baseline & 2.33 & 36.67 & 33.33 & 33.33 & 9.81 & 25.00 & 21.47 \\
& CANDID & 2.33 & 33.33 & 50.00 & 29.17 & 24.09 & 25.00 & \textbf{24.71} {\color{gray}(+3.24)} \\
\midrule

\multirow{2}{*}{Qwen3-VL-8B}
& Baseline & 0.00 & 20.00 & 16.67 & 12.50 & 14.29 & 0.00 & 10.39 \\
& CANDID & 0.00 & 23.33 & 20.83 & 12.50 & 14.29 & 0.00 & \textbf{11.69} {\color{gray}(+1.30)} \\
\bottomrule
\end{tabular}
}
\caption{Evaluation results on 154 WindowsAgentArena tasks with a step budget of $s=50$ and history length $h=4$. Values in parentheses next to domain names indicate the task count. Results are based on a single run; values in gray parentheses represent the absolute improvement over each model's baseline.}
\label{tab:waa-results}
\end{table*}

\subsection{Generalization to WindowsAgentArena}
\label{sec:waa-results}
We further evaluate CANDID on 154 WindowsAgentArena tasks to assess
cross-benchmark generalization. As shown in Table~\ref{tab:waa-results},
CANDID improves overall task accuracy from 21.47\% to 24.71\% for
Qwen3-VL-30B-A3B and from 10.39\% to 11.69\% for Qwen3-VL-8B. For
Qwen3-VL-30B-A3B, the largest gains occur in Windows System and Media \&
Video, although performance varies across domains. As the results are based on
a single run, they provide preliminary evidence of cross-benchmark
generalization.

\section{Baseline Implementations}
\label{sec:appendix-feedback-baselines}

This section describes three feedback-style baselines implemented for the comparison. Self-Guidance is our own lightweight prompt-based intervention. The Loop Breaker and Verifier are \textit{VLAA-GUI-style} implementations: they are inspired by the STOP and RECOVER components described in VLAA-GUI, but they are adapted to our PyAutoGUI action space and local-model setting rather than reproduced exactly.

\subsection{Self-Guidance}
\label{sec:appendix-guidance}

Self-Guidance runs a separate diagnostic prompt before normal action generation
and injects the resulting text into the next action prompt as \texttt{Guidance}.
The guidance model discusses progress, local loops, what to avoid, the next
direction, and whether task completion looks unsafe.

\paragraph{Prompt skeleton.}
The system prompt asks the model to diagnose the desktop trajectory:
\begin{center}
\setlength{\fboxsep}{6pt}
\fbox{%
\begin{minipage}{0.92\linewidth}
\small\raggedright
\textbf{System Prompt}\par
\smallskip
You are diagnosing progress in a desktop-computer task. Analyze the
instruction, recent screenshot-action trajectory, and current screenshot.
Produce short free-form guidance for the next action. Be concrete. Mention if
the agent appears stuck in a local loop. Mention what should not be repeated.
Mention whether task completion looks unsafe and should be avoided.
Do not output tool calls, XML tags, code, or executable UI actions.
\end{minipage}}
\end{center}
The user message contains the task instruction, recent screenshots and actions,
and the current screenshot. The requested output format is:
\begin{center}
\setlength{\fboxsep}{6pt}
\fbox{%
\begin{minipage}{0.92\linewidth}
\small\raggedright
\textbf{Requested Output Format}\par
\smallskip
\texttt{- progress so far}\\
\texttt{- suspected issue}\\
\texttt{- avoid repeating}\\
\texttt{- next direction}\\
\texttt{- completion note}
\end{minipage}}
\end{center}

\paragraph{Output use.}
The output is not parsed as a formal decision. If non-empty, it is prepended to
the next normal action-generation prompt as:
\begin{center}
\setlength{\fboxsep}{6pt}
\fbox{%
\begin{minipage}{0.92\linewidth}
\small\raggedright
\textbf{Control Block Appended to the User Prompt}\par
\smallskip
\texttt{Guidance:}\\
\texttt{<free-form guidance text>}
\end{minipage}}
\end{center}
The action model may ignore or follow the advice; no hard candidate filtering
or action constraint is applied.

\subsection{VLAA-GUI-Style Loop Breaker}
\label{sec:appendix-vlaa-loop}

The VLAA-GUI-style Loop Breaker is a reflection module that runs before action generation when previous responses exist. It follows the RECOVER idea from VLAA-GUI: inspect recent trajectory evidence, decide whether the agent is stuck in repetitive no-progress behavior, and emit a strategy-change signal. In our implementation, the module does not directly choose an action. Instead, when it returns a loop signal, it writes a textual loop-recovery constraint into the next action prompt.

\paragraph{Prompt skeleton.}
The system prompt is:
\begin{center}
\setlength{\fboxsep}{6pt}
\fbox{%
\begin{minipage}{0.92\linewidth}
\small\raggedright
\textbf{System Prompt}\par
\smallskip
You are a reflection agent for a desktop automation task. Analyze the recent trajectory and current screenshot for repetitive no-progress behavior. Focus only on loop recovery. Do not judge whether the whole task is complete. Return only a valid JSON object.
\end{minipage}}
\end{center}
At each step, the user message includes the latest assistant response, the parsed action, and optionally the current screenshot. The requested schema is:
\begin{center}
\setlength{\fboxsep}{6pt}
\fbox{%
\begin{minipage}{0.92\linewidth}
\small\raggedright
\textbf{Requested Output Schema}\par
\smallskip
\ttfamily
\{"progress\_signal": "what visibly changed, or no visible progress",\\
"loop\_detected": true/false,\\
"loop\_evidence": "short evidence",\\
"strategy\_signal": "KEEP or SWITCH",\\
"avoid\_actions": ["short action/target patterns to avoid"],\\
"escape\_direction": "short direction for changing approach"\}
\end{minipage}}
\end{center}

\paragraph{Output use.}
If \texttt{loop\_detected=true} or \texttt{strategy\_signal=SWITCH}, the module formats the result into a constraint block:
\begin{center}
\setlength{\fboxsep}{6pt}
\fbox{%
\begin{minipage}{0.92\linewidth}
\small\raggedright
\textbf{Control Block Appended to the User Prompt}\par
\smallskip
\texttt{Loop recovery constraint:}\\
\texttt{- strategy\_signal: SWITCH}\\
\texttt{- loop\_evidence: <evidence>}\\
\texttt{- avoid\_actions: <patterns>}\\
\texttt{- escape\_direction: <direction>}\\
\texttt{The next action must change approach and must not match
avoid\_actions.}
\end{minipage}}
\end{center}
This block is injected into the next action-generation prompt. The enforcement
mode is prompt-only: unlike our candidate-based loop recovery, the VLAA-GUI-style
Loop Breaker does not sample multiple executable candidates or remove repeated
actions through a hard filter.

\subsection{VLAA-GUI-Style Completion Verifier}
\label{sec:appendix-vlaa-verifier}

The VLAA-GUI-style Verifier is activated only when the normal action model proposes \texttt{DONE}. It follows the STOP idea from VLAA-GUI: accept task termination only when the current screenshot or recent trajectory provides direct evidence that every task requirement is satisfied. Our implementation is intentionally conservative and rejects ambiguous completion claims.

\paragraph{Prompt skeleton.}
The system prompt is:
\begin{center}
\setlength{\fboxsep}{6pt}
\fbox{%
\begin{minipage}{0.92\linewidth}
\small\raggedright
\textbf{System Prompt}\par
\smallskip
You are an extremely strict task completion verifier for desktop automation. Decide whether the task is already fully completed right now. Mark completion only if every requirement has direct, unambiguous evidence in the current screenshot or recent trajectory text. Do not assume hidden state or intent. For side effects such as saving, exporting, sending, creating, deleting, or changing settings, require visible or trajectory evidence that the side effect was committed. Prefer false negatives over false positives. Return only a valid JSON object.
\end{minipage}}
\end{center}
The user message includes the task instruction, recent trajectory text, the candidate final response, and the parsed termination action. The requested schema is:
\begin{center}
\setlength{\fboxsep}{6pt}
\fbox{%
\begin{minipage}{0.92\linewidth}
\small\raggedright
\textbf{Requested Output Schema}\par
\smallskip
\ttfamily
\{"complete": true/false,\\
"reason": "short evidence-based reason",\\
"missing\_steps": "what evidence/action is still needed, or empty if
complete"\}
\end{minipage}}
\end{center}

\paragraph{Output use.}
If the Verifier returns \texttt{complete=false}, or if it returns
\texttt{complete=true} while still mentioning uncertainty or missing steps, the agent rejects the termination. The current action is converted to a wait-style continuation, and the next prompt receives feedback of the form:
\begin{center}
\setlength{\fboxsep}{6pt}
\fbox{%
\begin{minipage}{0.92\linewidth}
\small\raggedright
\textbf{Control Block Appended to the User Prompt}\par
\smallskip
\texttt{Completion verifier rejected the previous terminate action:}\\
\texttt{- reason: <reason>}\\
\texttt{- missing\_steps: <missing evidence or action>}\\
\texttt{Continue the task. Do not terminate again until the missing evidence is visible.}
\end{minipage}}
\end{center}
Thus, the VLAA-GUI-style Verifier is a feedback mechanism rather than a candidate-selection mechanism: it can block a \texttt{DONE} action and provide textual correction, but it does not itself select among concrete continuation actions.

\section{Case Study}
\label{sec:appendix-case-study}

\subsection{Loop-Recovery Candidate Selection}

Fig.~\ref{fig:appendix-loop-recovery-example} shows a concrete loop-recovery event from a Qwen3-VL 8B trajectory. The agent repeatedly pressed \texttt{Enter} while reading the Miniconda license text; the loop detector identified this as an exact repeated key action, the confirmation gate validated the no-progress pattern, and candidate filtering selected a different continuation action.

\begin{figure*}[t]
\centering
\fbox{%
\begin{minipage}{0.95\textwidth}
\small
\textbf{Loop-recovery example from a Qwen3-VL 8B trajectory.}
The task is a multi-application setup task involving Miniconda installation.
At step 20, the baseline action proposal would have continued pressing
\texttt{Enter} while the terminal was displaying the license text.

\noindent\textbf{Recent executed actions.}
\texttt{press('enter')} $\rightarrow$
\texttt{press('enter')} $\rightarrow$
\texttt{press('enter')} $\rightarrow$
\texttt{press('enter')} $\rightarrow$
\texttt{press('enter')}

\noindent\textbf{Raw proposed action.}
``Press the Enter key to continue reading the Miniconda license agreement.''
Parsed code: \texttt{pyautogui.press('enter')}.

\noindent\textbf{Rule trigger.}
The detector fires with
\texttt{repeated\_exact\_text\_or\_key\_action}: the proposed action type is
\texttt{key}, the normalized action is exactly
\texttt{pyautogui.press('enter')}, and the same action appears 5 times in the
recent window. The cyclic detector also observes a repeated key pattern with
repeat count 3.

\noindent\textbf{Model confirmation.}
The confirmation gate marks the trajectory as stagnant:
\textit{``License text scrolling, no installation progress''}; it sets \texttt{loop\_detected=true} and \texttt{strategy\_signal=SWITCH}, with the escape direction: \textit{``Press 'n' or 'no' to skip license and proceed to installation.''}

\noindent\textbf{Candidate filtering and selected recovery action.}
Four recovery candidates are sampled. Three candidates are blocked because
they are still \texttt{key} actions without a distinguishable target under the
blocked action type. The only selectable candidate changes the action type to
\texttt{type} and executes \texttt{pyautogui.typewrite('n')}. Since it is the
sole valid non-blocked candidate, the selector is skipped and this action is
executed directly.
\end{minipage}}
\caption{Concrete loop-recovery event of a Qwen3-VL 8B trajectory.
The original action would have repeated \texttt{Enter}; loop recovery instead
executes \texttt{pyautogui.typewrite('n')} after rule detection, model
confirmation, and candidate filtering.}
\label{fig:appendix-loop-recovery-example}
\end{figure*}

\subsection{Unanimous Done Gating}

Fig.~\ref{fig:appendix-done-selection-example} shows a concrete DONE-selection event from an OpenCUA-7B trajectory. The initial action proposed successful termination even though the email composition window was still open. Additional samples disagreed with termination and proposed clicking the visible Send button, so the unanimous policy rejected \texttt{DONE} and forced a concrete continuation action.

\begin{figure*}[t]
\centering
\fbox{%
\begin{minipage}{0.95\textwidth}
\small
\textbf{DONE-selection example from an OpenCUA-7B trajectory.}
The task is a LibreOffice Writer sharing task. At step 10, the document is
attached to an email composition window, but the message has not yet been sent.

\noindent\textbf{Initial termination proposal.}
The baseline proposal is parsed as \texttt{DONE}: it claims that the document has been shared successfully and calls \texttt{computer.terminate(status='success')}. However, the visible state still contains the email composition window and the Send button, indicating that a final send action is still required.

\noindent\textbf{Candidate disagreement.}
The DONE gating module samples four additional candidates from the same context. The resulting candidate set contains five candidates in total: the original \texttt{DONE}, one additional \texttt{DONE}, and three valid non-\texttt{DONE} continuations. Because valid continuation actions exist, the unanimous policy rejects termination and excludes all DONE candidates before selection.

\noindent\textbf{Selector decision.}
The remaining candidates all attempt the missing final operation: clicking the Send button in the email composition toolbar. The selector chooses candidate 1 with the reason that prior Send-click attempts likely missed the correct coordinates and that clicking the visible Send button is the final required step. The selected continuation executes \texttt{pyautogui.click(152, 144)}, redirecting the original DONE decision to a concrete non-DONE action.
\end{minipage}}
\caption{Concrete unanimous DONE-gating event of an OpenCUA-7B trajectory. The initial proposal terminates with success, but sampled candidates reveal a valid non-DONE continuation, so DONE candidates are excluded and the agent clicks the Send button instead.}
\label{fig:appendix-done-selection-example}
\end{figure*}






\end{document}